\documentclass{article}

\usepackage{PRIMEarxiv}

\usepackage[utf8]{inputenc} 
\usepackage[T1]{fontenc}    
\usepackage[numbers]{natbib}
\usepackage{hyperref}       
\usepackage{url}            
\usepackage{booktabs}       
\usepackage{amsfonts}       
\usepackage{nicefrac}       
\usepackage{microtype}      
\usepackage{lipsum}
\usepackage{fancyhdr}       
\usepackage{graphicx}       
\graphicspath{{media/}}     

\usepackage{hyperref}
\usepackage{url}

\usepackage{enumitem}
\usepackage{amsmath,amsthm,amssymb}
\usepackage{algorithm}
\usepackage{algorithmic}
\usepackage{graphicx}
\usepackage{colortbl}
\usepackage{dsfont}
\usepackage{float} 
\usepackage{authblk}
\usepackage{caption}
\usepackage{subfigure}
\usepackage{booktabs}
\usepackage{makecell}

\renewcommand{\algorithmicrequire}{ \textbf{Input:}} 
\renewcommand{\algorithmicensure}{ \textbf{Output:}} 

\usepackage{microtype}
\usepackage{graphicx}
\usepackage{booktabs} 

\usepackage{hyperref}

\title{On the Effectiveness of Adversarial Training against Backdoor Attacks

}

\author{
Yinghua Gao\textsuperscript{\rm 1}\thanks{Equal contribution. }$\;\,$, 
Dongxian Wu\textsuperscript{\rm 2}$^{*}$, 
Jingfeng Zhang\textsuperscript{\rm 3}\thanks{Correspondence to: Jingfeng Zhang <jingfeng.zhang@riken.jp> and Shu-Tao Xia <xiast@sz.tsinghua.edu.cn>}$\;\,$ ,
Guanhao Gan\textsuperscript{\rm 1},
Shu-Tao Xia\textsuperscript{\rm 1}$^{\dag}$ ,
Gang Niu\textsuperscript{\rm 3},
Masashi Sugiyama\textsuperscript{\rm 3 2}\\
\textsuperscript{\rm 1} Tsinghua University\\
\textsuperscript{\rm 2}  The University of Tokyo\\ 
\textsuperscript{\rm 3}  RIKEN Center for Advanced Intelligence Project (AIP)\\ 
}

\begin{document}
\maketitle

\begin{abstract}
DNNs' demand for massive data forces practitioners to collect data from the Internet without careful check due to the unacceptable cost, which brings potential risks of backdoor attacks. A backdoored model always predicts a target class in the presence of a predefined trigger pattern, which can be easily realized via poisoning a small amount of data. In general, adversarial training is believed to defend against backdoor attacks since it helps models to keep their prediction unchanged even if we perturb the input image (as long as within a feasible range). Unfortunately, few previous studies succeed in doing so. To explore whether adversarial training could defend against backdoor attacks or not, we conduct extensive experiments across different threat models and perturbation budgets, and find the threat model in adversarial training matters. For instance, adversarial training with spatial adversarial examples provides notable robustness against commonly-used patch-based backdoor attacks. We further propose a hybrid strategy which provides satisfactory robustness across different backdoor attacks.
\end{abstract}


\section{Introduction}

As deep neural networks (DNNs) require massive amounts of data, practitioners have to crawl images and labels from websites, which brings potential risks such as backdoor attacks \citep{gu2017badnets,chen2017targeted, li2020backdoor, goldblum2020dataset}. Specifically, an adversary could easily backdoor a classifier via poisoning a small amount of data, \textit{i.e.}, patching a trigger on a few training data and (sometimes) relabeling them as a predefined class. As a result, the backdoored model would always misclassify a test image into a target class in the presence of the trigger pattern, while it behaves normally on benign images. For example, it has been illustrated that one could use a sticker as the trigger to mislead a road sign classifier to identify `stop' signs to `speed limited' signs \citep{gu2017badnets}. Since backdoor attacks bring remarkable threat to safety-critical applications such as autonomous driving \citep{ding2019trojan} and smart healthcare \citep{ali2020smart}, it is urgent to defend against such attacks during training \citep{hong2020effectiveness,weber2020rab,borgnia2021dp}. 

Recently, Adversarial Training (AT) \citep{goodfellow2014explaining,madry2018towards} becomes a popular method in trustworthy machine learning, not only because it provides empirical robustness against adversarial examples \citep{szegedy2014intriguing}, but also because of the benefits it can provide in terms of transfer learning \citep{salman2020adversarially}, clustering \citep{bai2021clustering}, interpretability \citep{tsipras2018robustness}, and generalization \citep{xie2020adversarial}. AT formulates a minimax optimization in which we want the trained classifier
to maintain the prediction even if the input image is perturbed, that is,
\begin{equation}
\min_{\theta}\sum_{i=1}^n \max_{x_i^\prime \in \mathcal{B}(x_i)}   \ell (f_\theta(x_i^\prime), y_i),
\label{eqn:adv_train}
\end{equation}
where $n$ is the number of training examples, $x_i^\prime$ is the adversarial example (the worst case) within a feasible range $\mathcal{B}(x_i)$, $f_{\theta}(\cdot)$ is the DNN with parameters $\theta$, $\ell(\cdot)$ is the standard classification loss (\textit{e.g.}, the cross-entropy loss). We also term the feasible range as the threat model in AT and a commonly-used one is the $L_\infty$-norm ball ($\Vert x_i^\prime - x_i \Vert_\infty \leq \epsilon$), that is, the perturbation on any single pixel cannot exceed $\epsilon$.
AT is believed to provide robustness against backdoor attacks because an adversarially trained model could keep the prediction unchanged when the input image is perturbed (\textit{e.g.}, patched by a trigger pattern). Unfortunately, previous studies only achieved unsatisfactory performance \citep{geiping2021doesn} or even claimed that AT strengthens the backdoor vulnerability \citep{weng2020trade}. Therefore, we explore a question of \textit{whether AT could effectively defend against backdoor attacks}.

To answer the above question, we study how different settings in AT affect backdoor robustness, including the threat models and perturbation budgets. After conducting extensive experiments across varying backdoor scenarios (poisoning types, trigger shapes and sizes), we find that the threat model in AT matters in backdoor robustness. In particular, for the commonly-used patch-based  backdoor attack (\textit{i.e.}, the trigger pattern is a predefined patch) \citep{gu2017badnets}, AT with spatial adversarial examples (spatial AT) \citep{xiao2018spatially} surprisingly provides significant robustness, while AT with $L_p$ adversarial examples ($L_p$ AT) fails in it. In addition, we did not observe that the backdoor robustness deteriorates in the spatial AT as the perturbation budget increases, unlike the phenomenon in $L_p$ AT \citep{weng2020trade}. Meanwhile, whole-image backdoor attacks \citep{chen2017targeted} could easily escape the defense from spatial AT, while still being mitigated by $L_p$ AT.
 Inspired by these findings, we propose a hybrid strategy to help practitioners effectively tackle with backdoor attacks. Our work is related to a recent work \citep{tao2021better} which attempts to prevent delusive
attacks (usually indiscriminate) with AT. However, our findings are more general since we explore the possibility of AT against both discriminate and indiscriminate backdoor attacks. We compare with recent state-of-the-art backdoor defense methods and discuss the advantages when we apply AT against backdoor attacks such as no demand for extra clean data like the defense after training \citep{liu2018fine, li2021neural} or no need to isolate poisoned and clean samples like Li \textit{et al.} \citep{li2021anti}. Moreover, we consider the scenario that the adversary knows our AT method and experiment with adaptive attack.    
In conclusion, our main contributions of the paper can be summarized as follows:

\begin{itemize}[leftmargin=*]
\item We provide a systematic evaluation on the backdoor attacks with AT. 

\item We identify effective adversarial perturbations which mitigate specific backdoor attacks, and propose a hybrid strategy to tackle with backdoor attacks.

\item Through extensive experiments, including the comparison with recent baseline defense methods and adaptive attack, we demonstrate the effectiveness of adversarial training against backdoor attacks.
\end{itemize}

\section{Background and Preliminary}

AT is varied across the threat models with different defensive effects against backdoor attacks. We first introduce different threat models in AT, and then discuss different types of backdoor attacks.

\subsection{AT with Different Threat Models}
\label{sec:back_at}

AT can be categorized according to the definition of $\mathcal{B}(x_i)$ as well as how to solve the inner maximization in Equation (\ref{eqn:adv_train}). This paper  mainly considers the following types of AT:

{\bf $L_p$
AT} \citep{madry2018towards}. $L_p$ 
AT is the most common method and has been extensively studied \citep{madry2018towards,zhang2019theoretically,wang2019improving,zhang2020attacks, wu2020adversarial, zhang2021geometry, bai2021improving}. In this threat model, we require the perturbation is not larger than $\epsilon$ in $L_p$-norm, \textit{i.e.}, $\mathcal{B}(x_i)=\left\{x_i^\prime| \Vert x_i^\prime - x_i \Vert_p \leq \epsilon \right\} $. Usually, we adopt the Projected Gradient Descent (PGD) method to solve the inner maximization as suggested in Madry \textit{et al.} \citep{madry2018towards}. This paper considers $p=\infty$ and $p=2$ that are commonly used in previous research.

{\bf Spatial AT} \citep{xiao2018spatially}. To create more distinguishable adversarial examples, Xiao \textit{et al.} \citep{xiao2018spatially} proposed spatially transformed examples by changing the positions of pixels rather than directly modifying pixel values. In spatial AT, the inner objective is a sum of a classification loss and a spatial movement loss. In our work, a slight difference with  Xiao \textit{et al.} \citep{xiao2018spatially} is that we solve the inner maximization with the first order optimization rather than L-BFGS solver \citep{liu1989limited} in the original paper as L-BFGS solver can not enjoy the GPU acceleration.

{\bf Perceptual AT} \citep{laidlaw2020perceptual}.  To better correlate with human's perceptibility of adversarial examples, Laidlaw \textit{et al.} \citep{laidlaw2020perceptual} proposed neural perceptual threat models and utilized Learned Perceptual Image Patch Similarity (LPIPS) \citep{zhang2018unreasonable} as the surrogate for human vision: $d(x_i^\prime, x_i)=\Vert \phi(x_i^\prime)-\phi(x_i) \Vert_2$ and $\mathcal{B}(x_i)=\left\{x_i^\prime | d(x_i^\prime, x_i)\le\epsilon\right\}$, where $\phi(x_i)$ denotes the flattened
internal activation vector generated by a specific network \citep{zhang2018unreasonable}. We use the Lagrangian relaxation to solve the inner maximization and the self-bounded Perceptual AT which means the same network is used for training models and calculating the LPIPS distance  simultaneously. 

\subsection{Backdoor Attacks}
\label{sec:back_back}

 \begin{figure*}[!t]
  \centering
  \includegraphics[scale=0.6]{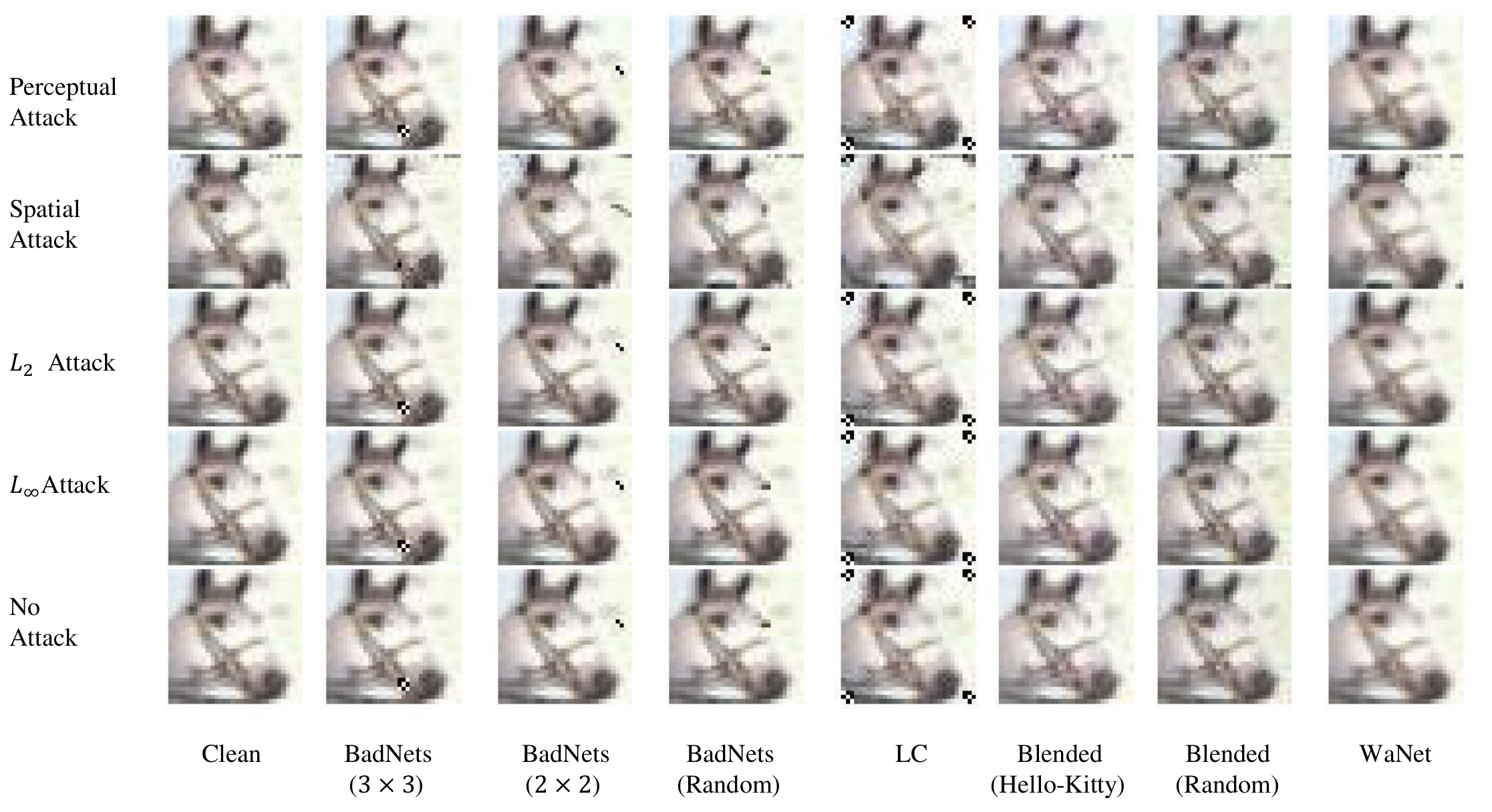}
  \caption{Illustrations of poisoned samples with different backdoor attacks and the adversarial examples generated by different adversarial attacks.}
  \label{fig:adv_back}
  \end{figure*}

In backdoor attacks, an adversary can poison a fraction of training data via attaching a predefined trigger pattern and relabeling them as target labels (dirty-label setting) or only poisoning the samples in the target class (clean label setting). After training, a backdoored model can be controlled to predict the predefined target label whenever the trigger patterns appear on the image. According to the trigger form, we divide the backdoor attacks into the patch-based attack (trigger is a local patch) and the whole-image attack (trigger is a perturbation over the entire image). We introduce four representative attacks as follows:


{\bf BadNets} \citep{gu2017badnets}. The simplest way is to patch a predefined pattern (\textit{e.g.}, a checkerboard) on an image. In such a case, the triggered sample $\Tilde{x}$ can be calculated as $\Tilde{x}=({\bf 1}-m)\odot x +m\odot t$, where $\odot$ denotes the element-wise multiplication, $x\in \mathbb{R}^d$ is the benign sample, $t\in \mathbb{R}^d$
is the predefined trigger pattern, ${\bf 1}$ is a $d$-dimensional all-one mask, $m\in \left\{0,1\right\}^d$ is a binary mask that determines the trigger injecting region. We consider a $3\times 3$ checkerboard trigger, $2\times 2$ checkerboard trigger and $2\times 2$ random trigger in our experiments. 

{\bf Label Consistent (LC) Attack} \citep{turner2019label}. To boost the performance of BadNets under the clean label setting, Turner \textit{et al.} \citep{turner2019label} proposed to add $L_p$
adversarial perturbations to the poisoned samples with an independently trained model. Specifically, we use a four-corner trigger as suggested in Turner \textit{et al.} \citep{turner2019label}.

{\bf Blended Attack} \citep{chen2017targeted}. A trigger patch (\textit{e.g.}, a checkerboard) in BadNets is easy to be detected. To achieve the stealthiness, Chen \textit{et al.} \citep{chen2017targeted} instead blended the benign image with a trigger pattern $t$, \textit{i.e.}, $\Tilde{x}=(1-\alpha)x+\alpha\cdot t$, where $\alpha \in (0, 1)$ is the transparency parameter concerned with the visibility of the trigger pattern. We consider a Hello-Kitty trigger and random trigger in our experiments.

{\bf WaNet} \citep{nguyen2021wanet}. To make the trigger unnoticeable, WaNet uses a smooth warping field to generate poisoned inputs. 

Among them, BadNets and LC are patch-based attacks and Blended and WaNet are whole-image attacks. LC is a clean label attack and the others are dirty label attacks. We illustrate the poisoned samples and the adversarial examples generated by different adversarial attacks in Figure \ref{fig:adv_back}.  


\subsection{Backdoor Defense}

Recently, numerous studies have been proposed to defend against backdoor attacks, including pruning \citep{liu2018fine,wu2021adversarial}, fine-tuning \citep{wang2019neural,zhao2020bridging}
, and distillation \citep{li2021neural}. Unfortunately, majority of them assume there is a small set of clean data, and repair the backdoored model after training. Different from them, this paper focuses on the defense during training, \textit{i.e.}, training a benign model from poisoned data. This paper considers two representative methods as follows:

{\bf  Differentially Private Stochastic Gradient Descent (DPSGD)} \citep{hong2020effectiveness}. The authors observed the $L_2$ 
norm difference of clean gradients in the presence of poisons and  proposed DPSGD, which clips and perturbs individual gradients during training to obtain privacy guarantees.

{\bf Anti-Backdoor Learning (ABL)} \citep{li2021anti}. ABL divides the standard training into two stages: the first stage to
identify the poisoned sample candidates with the amplified difference in the training loss and the second stage to unlearn the backdoor behavior with gradient ascent.

\section{Evaluation of Backdoor Vulnerability under AT }
\label{section-revisit}





Prior works achieved unsatisfactory robustness against backdoor attacks \citep{geiping2021doesn}. In addition, Weng \textit{et al.} \citep{weng2020trade} even have discovered that $L_{p}$ AT will strengthen the trigger memorization and hence weaken the model's backdoor robustness. 
Here we conduct extensive experiments to explore how AT impacts backdoor robustness, and answer following questions: Can vanilla AT (with a suitable threat model) address backdoor vulnerability?  Does AT always weaken the model's backdoor robustness across various threat models?

{\bf Experimental Settings}. \textit{1) Datasets}. We used CIFAR-10 and CIFAR-100 in our experiments. \textit{2) Threat models}. We considered four threat models ($L_\infty$, $L_2$, Spatial, and Perceptual) in Section \ref{sec:back_at}. Specifically, $L_\infty$ and $L_2$
AT were implemented based on the advertorch toolbox \citep{ding2019advertorch}. The perturbation budget ranges for $L_{\infty}$ AT was from $4/255$ to $16/255$, and the budget for $L_2$ AT was $64/255$ to $512/255$ respectively.
For spatial attacks, the maximal difference between adversarial and identity transformation ranged from 0.025 to $0.1$. For perceptual attacks, the perturbation budget ranged from $0.1$ to $0.5$.
\textit{3) Backdoor attacks}. We evaluated four backdoor attacks: BadNets with a $3\times 3$ checkerboard trigger, LC, Blended with a Hello-Kitty trigger, and WaNet in Section \ref{sec:back_back}. Following prior works \citep{weng2020trade}, we adopted the clean label setting for BadNets, which means we only poisoned the images belonging to the target class, while three other attacks were implemented based on the original papers. The poison rate  was $0.5\%$ for BadNets and LC, and $5\%$ for Blended and WaNet. For all attacks, class 2 was assigned as the target class. 
The settings for CIFAR-100 were similar and we leave the details in Appendix \ref{app:cifar100}. \textit{4) Training settings}. The normally and adversarially trained ResNet-18 \citep{he2016deep} 
models were obtained using an SGD optimizer for 100 epochs with the momentum $0.9$, the weight decay $5\times 10^{-4}$,
and the initial learning rate 0.1 which was divided by 10 at the 60-th and 90-th epochs. The common data augmentations such as random crop and random horizontal flip were used during training.

{\bf Evaluation Metrics}. In our experiments, we report the clean accuracy (ACC) which is the percentage of clean samples that are correctly classified and the attack success rate (ASR) which is the percentage of triggered samples that are predicted as the target label. 

\begin{figure*}[!htbp]
\centering
\subfigure[BadNets, $L_\infty $]{
\label{fig:badnets_linf}
\includegraphics[scale=0.22]{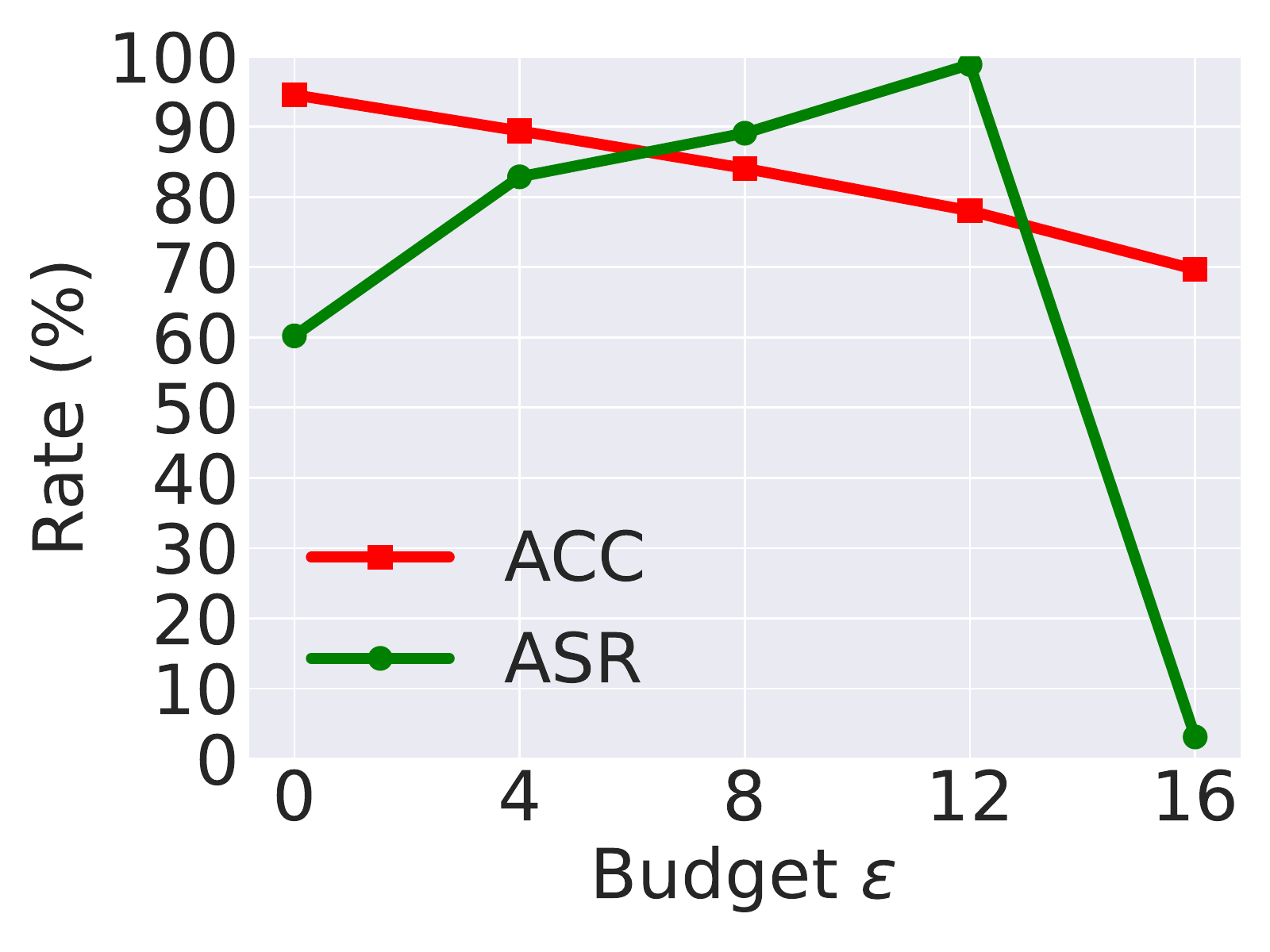}}
\subfigure[{LC, $L_\infty $}]{
\label{fig:lc_linf}
\includegraphics[scale=0.22]{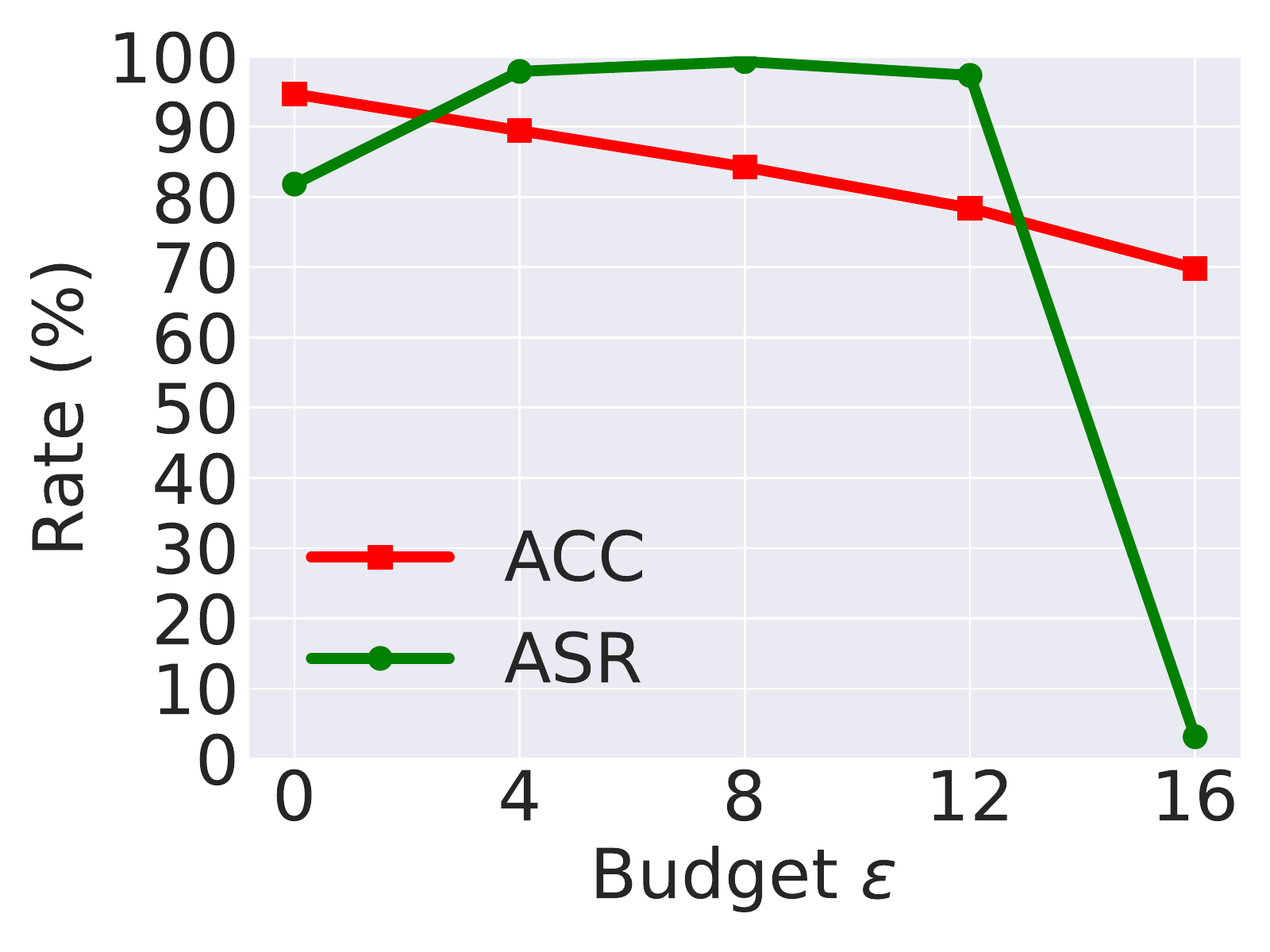}}
\subfigure[Blended, $L_\infty $]{
\label{fig:blended_linf}
\includegraphics[scale=0.22]{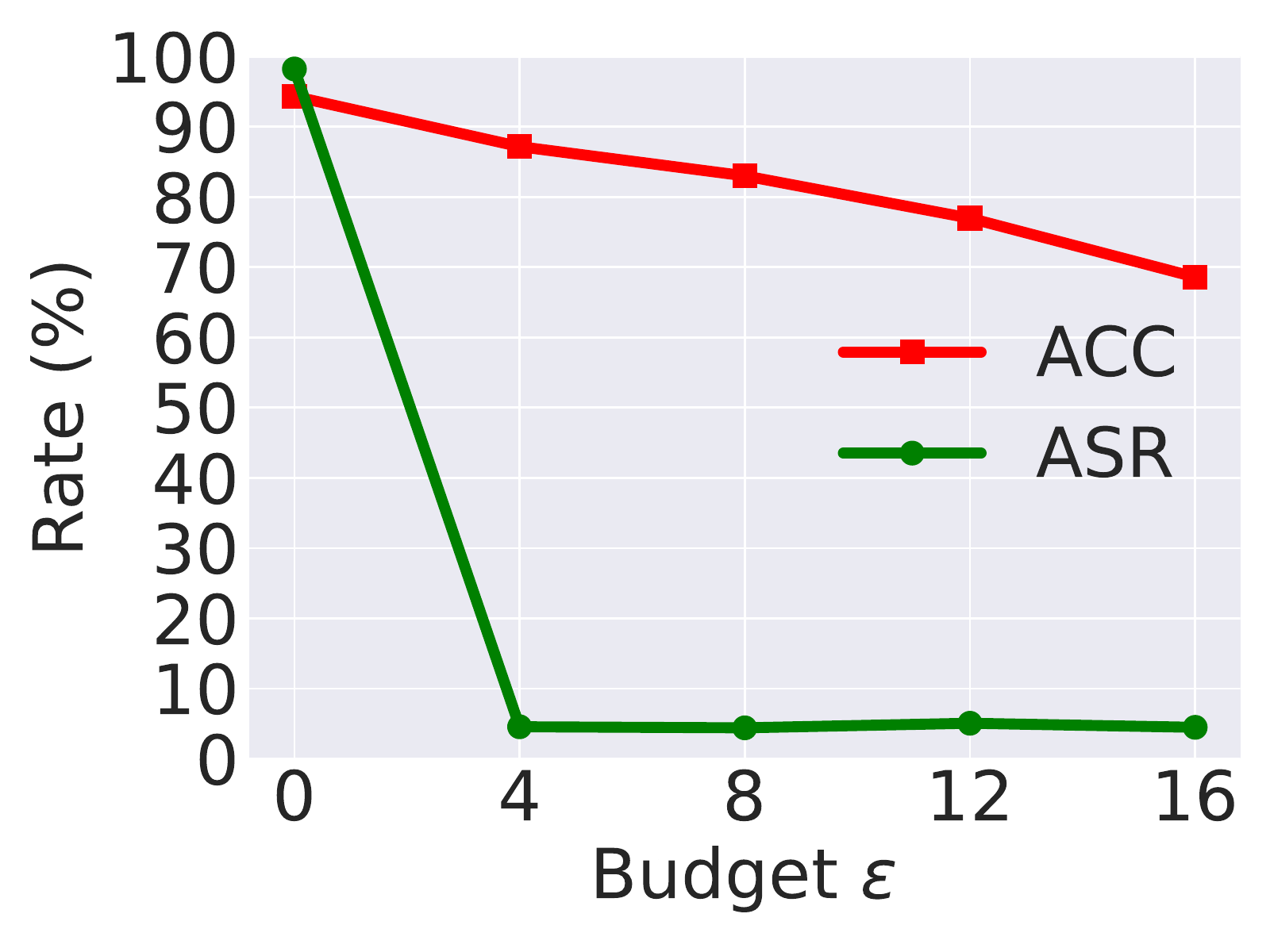}}
\subfigure[WaNet, $L_\infty $]{
\label{fig:wanet_linf}
\includegraphics[scale=0.22]{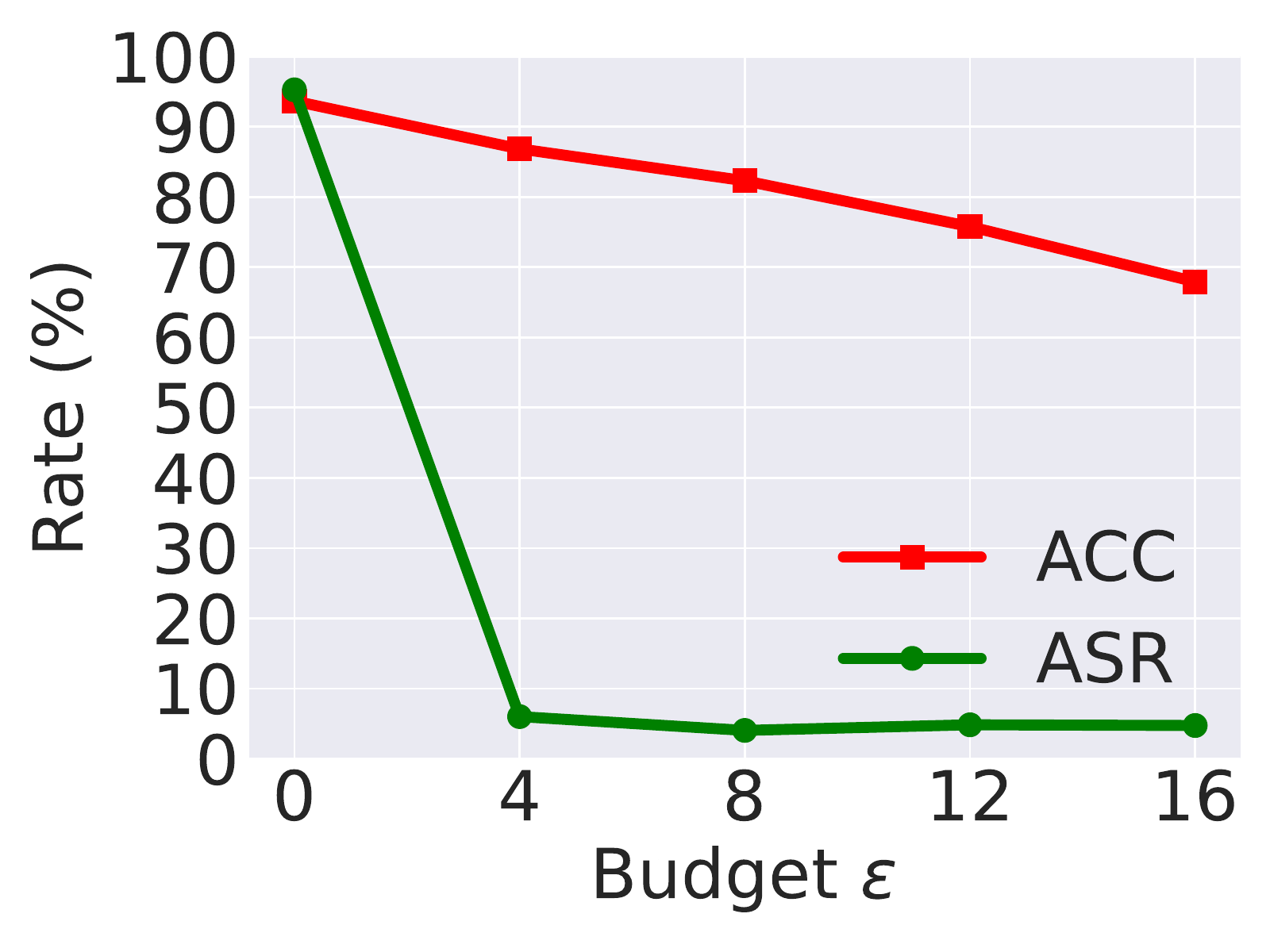}}

\subfigure[BadNets, $L_2 $]{
\label{fig:badnets_l2}
\includegraphics[scale=0.22]{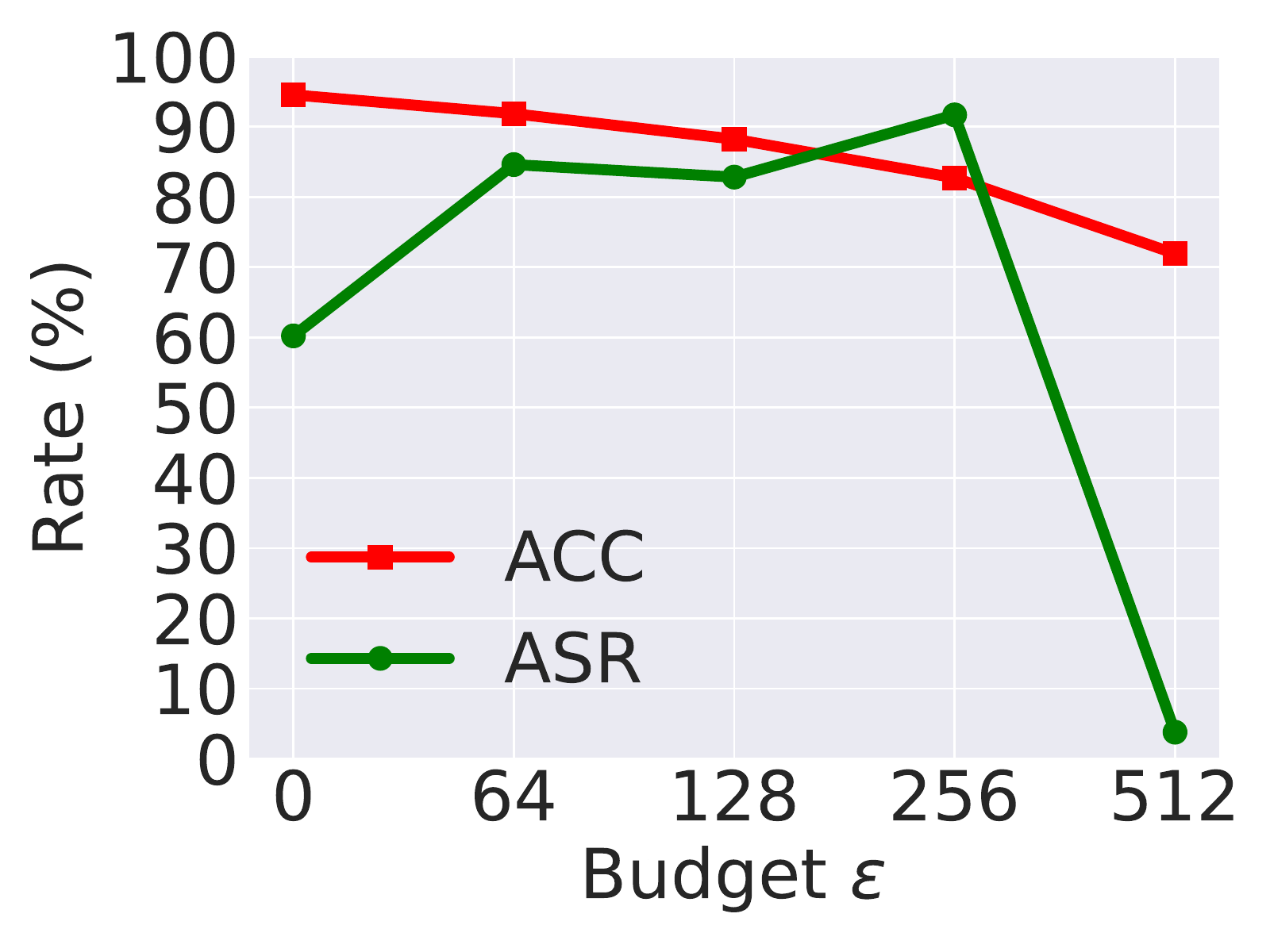}}
\subfigure[{LC, $L_2$}]{
\label{fig:lc_l2}
\includegraphics[scale=0.22]{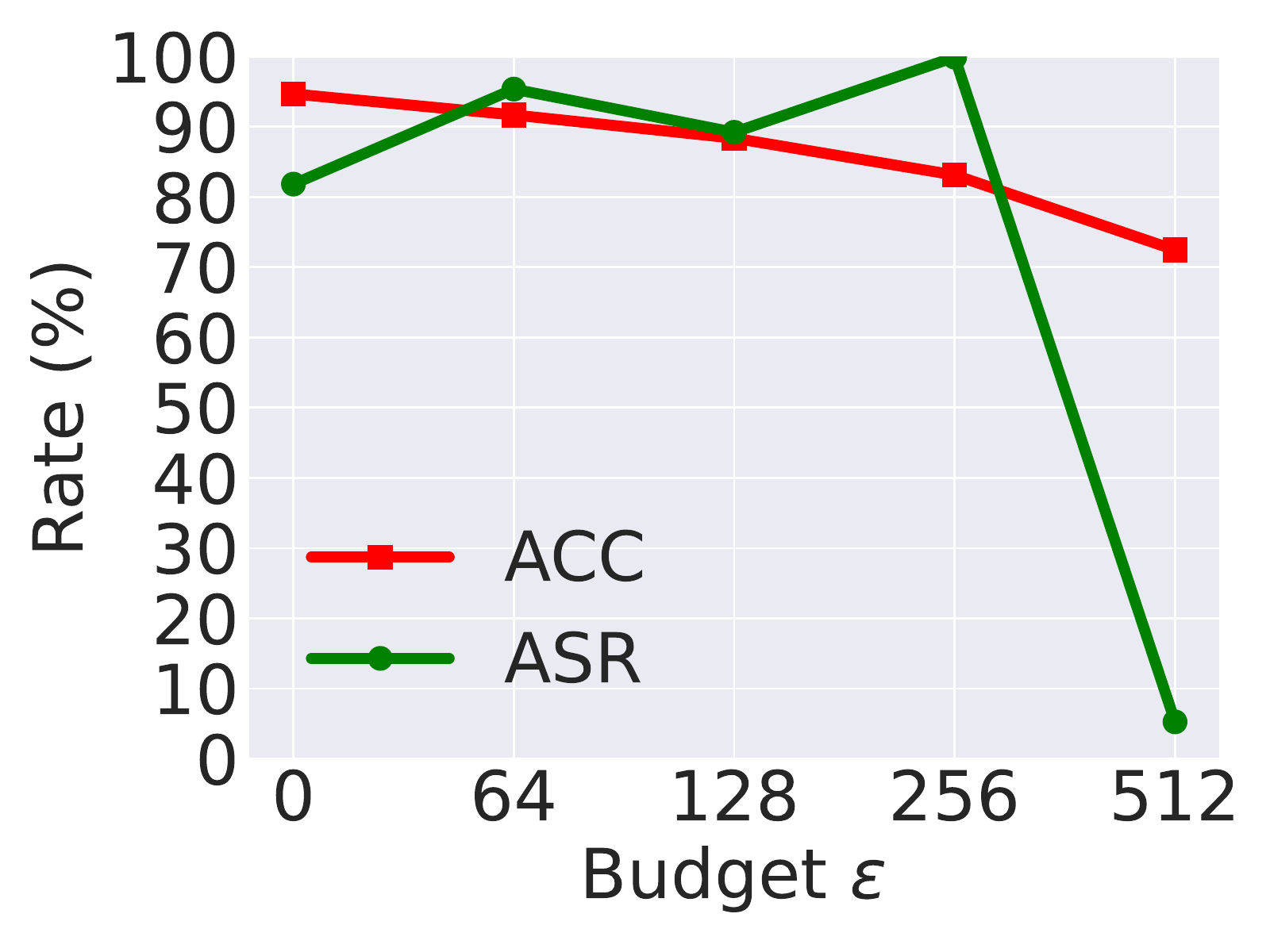}}
\subfigure[Blended, $L_2 $ ]{
\label{fig:blended_l2}
\includegraphics[scale=0.22]{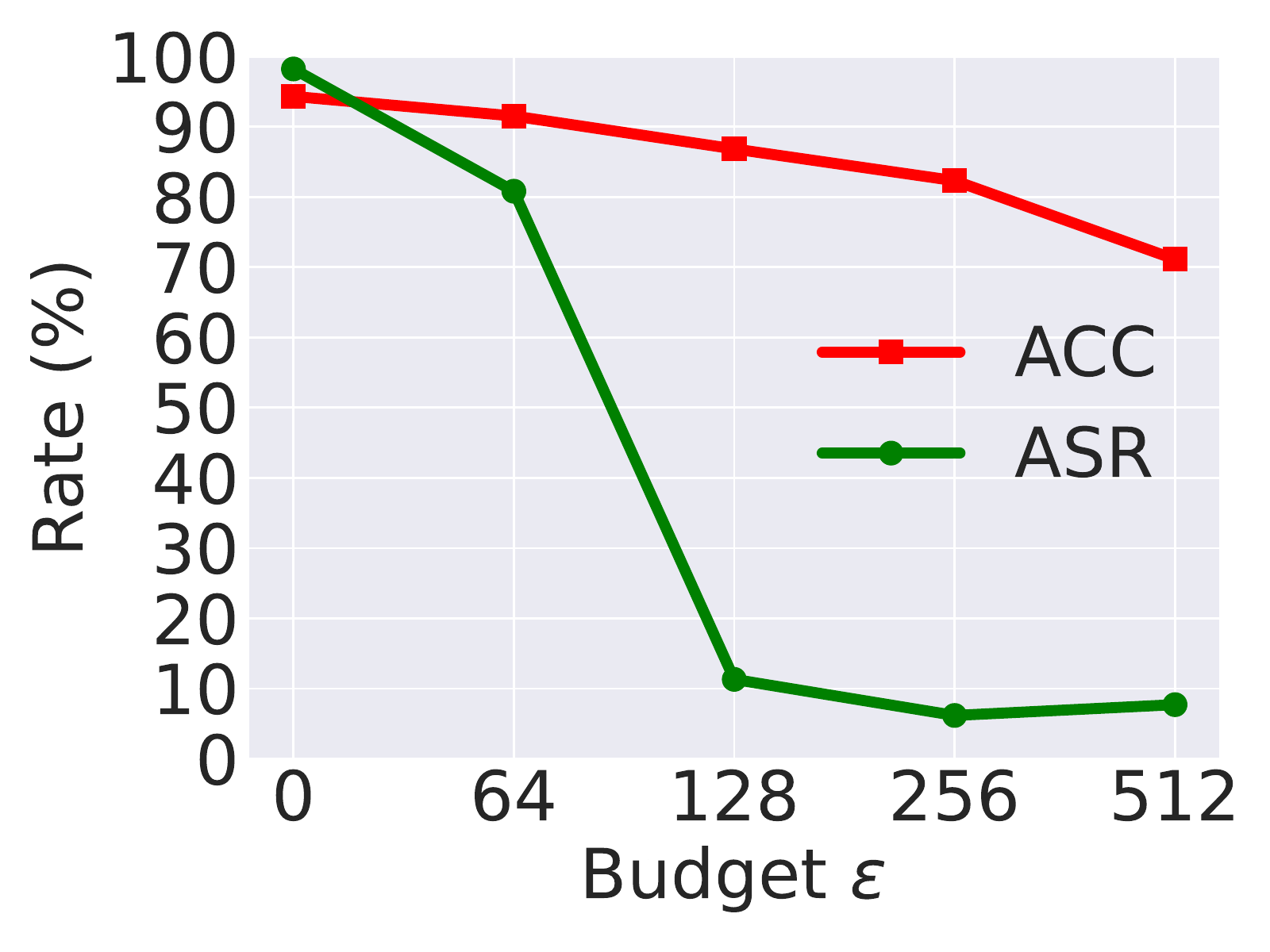}}
\subfigure[WaNet, $L_2$]{
\label{fig:wanet_l2}
\includegraphics[scale=0.22]{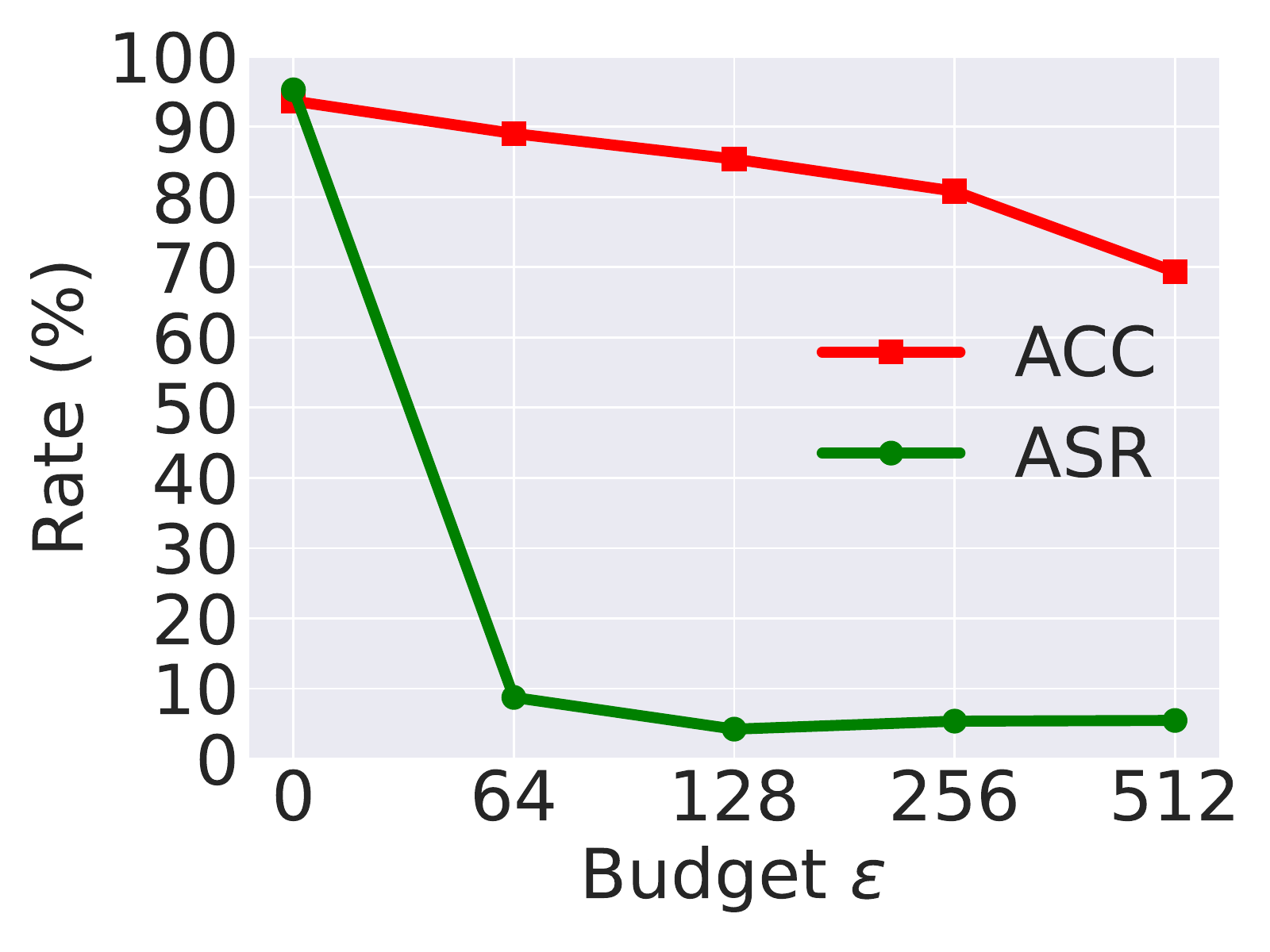}}

\subfigure[BadNets, Spatial]{
\label{fig:badnets_spatial}
\includegraphics[scale=0.22]{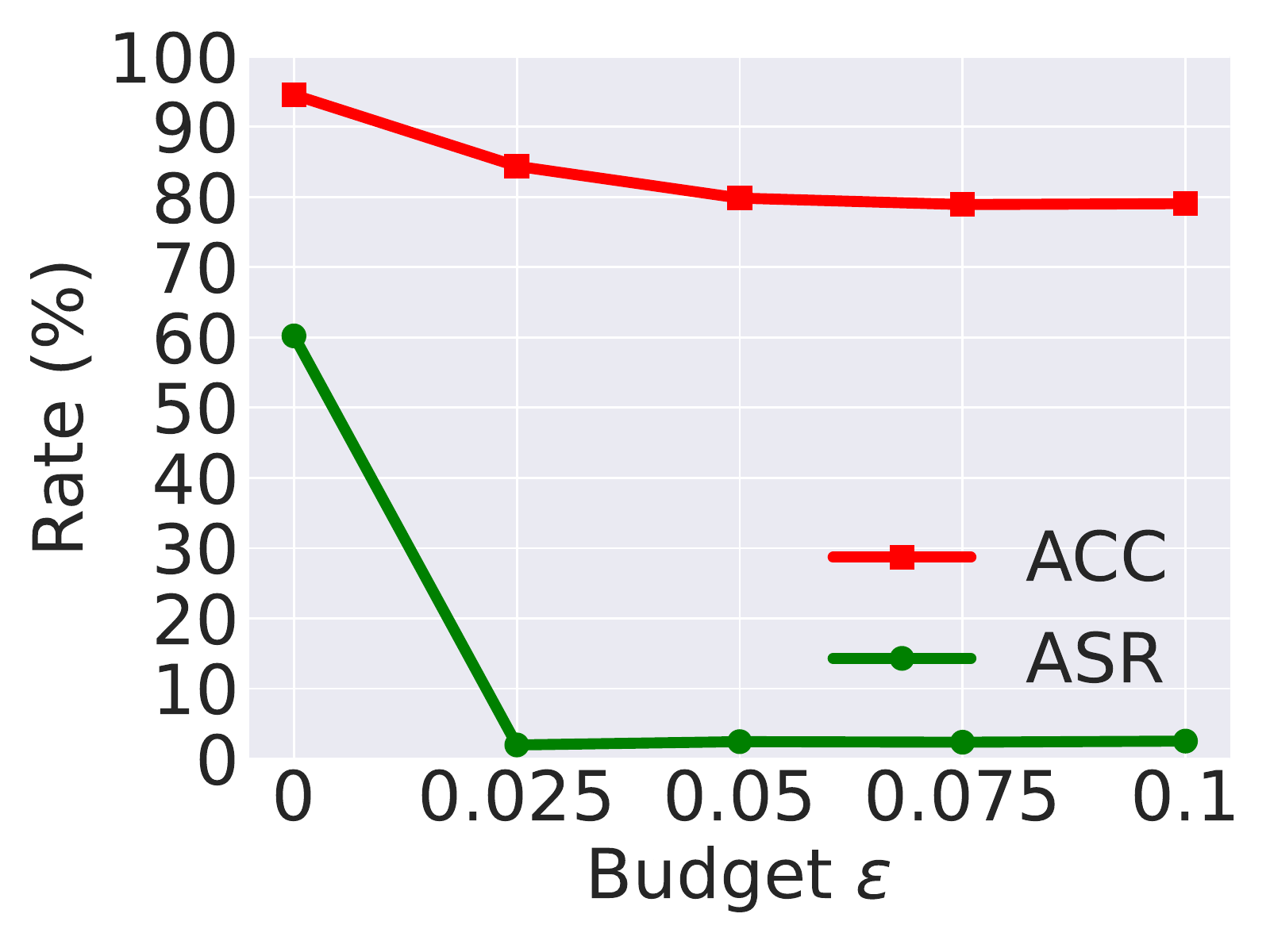}}
\subfigure[LC, Spatial]{
\label{fig:lc_spatial}
\includegraphics[scale=0.22]{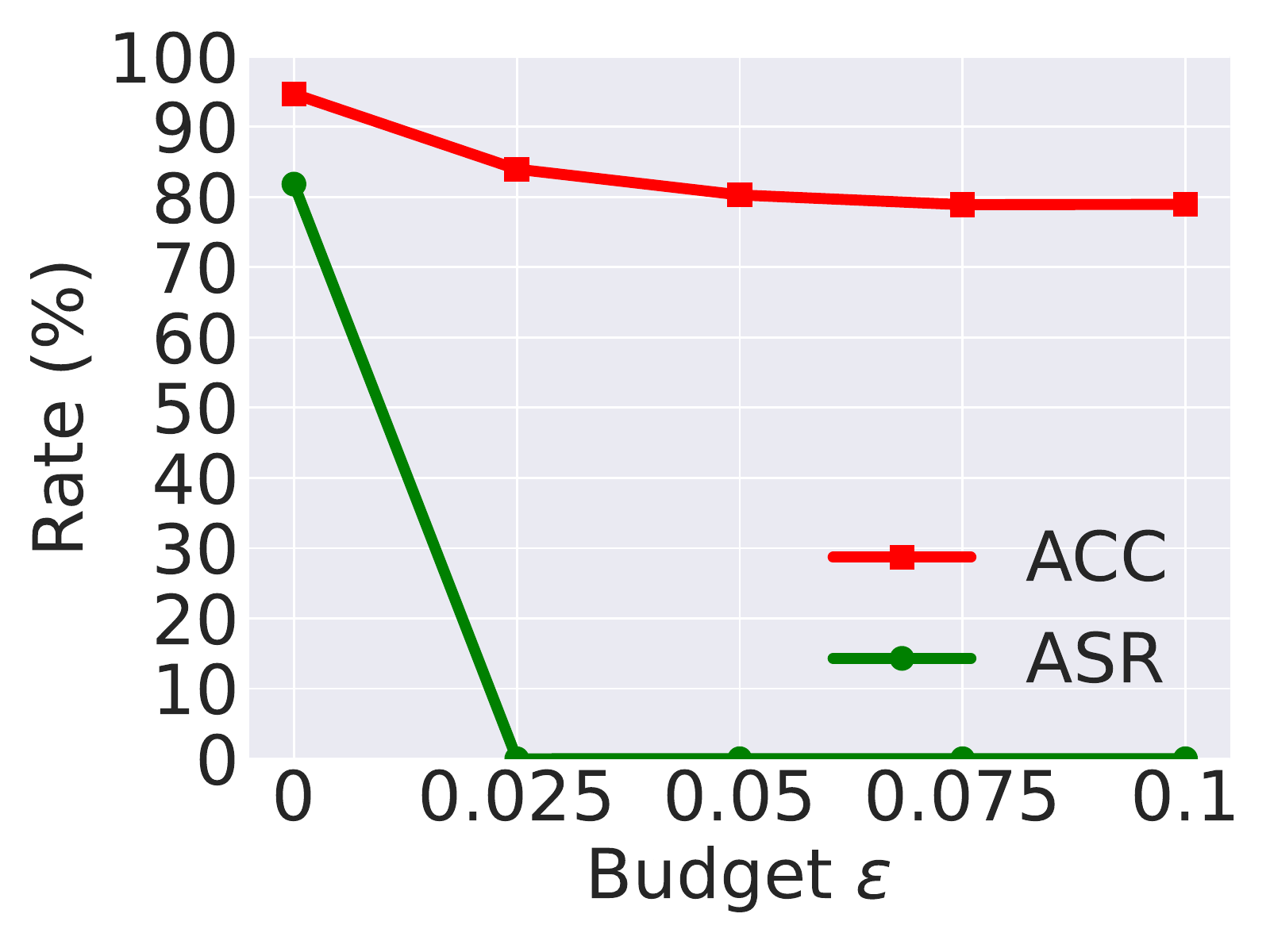}}
\subfigure[Blended, Spatial]{
\label{fig:blended_spatial}
\includegraphics[scale=0.22]{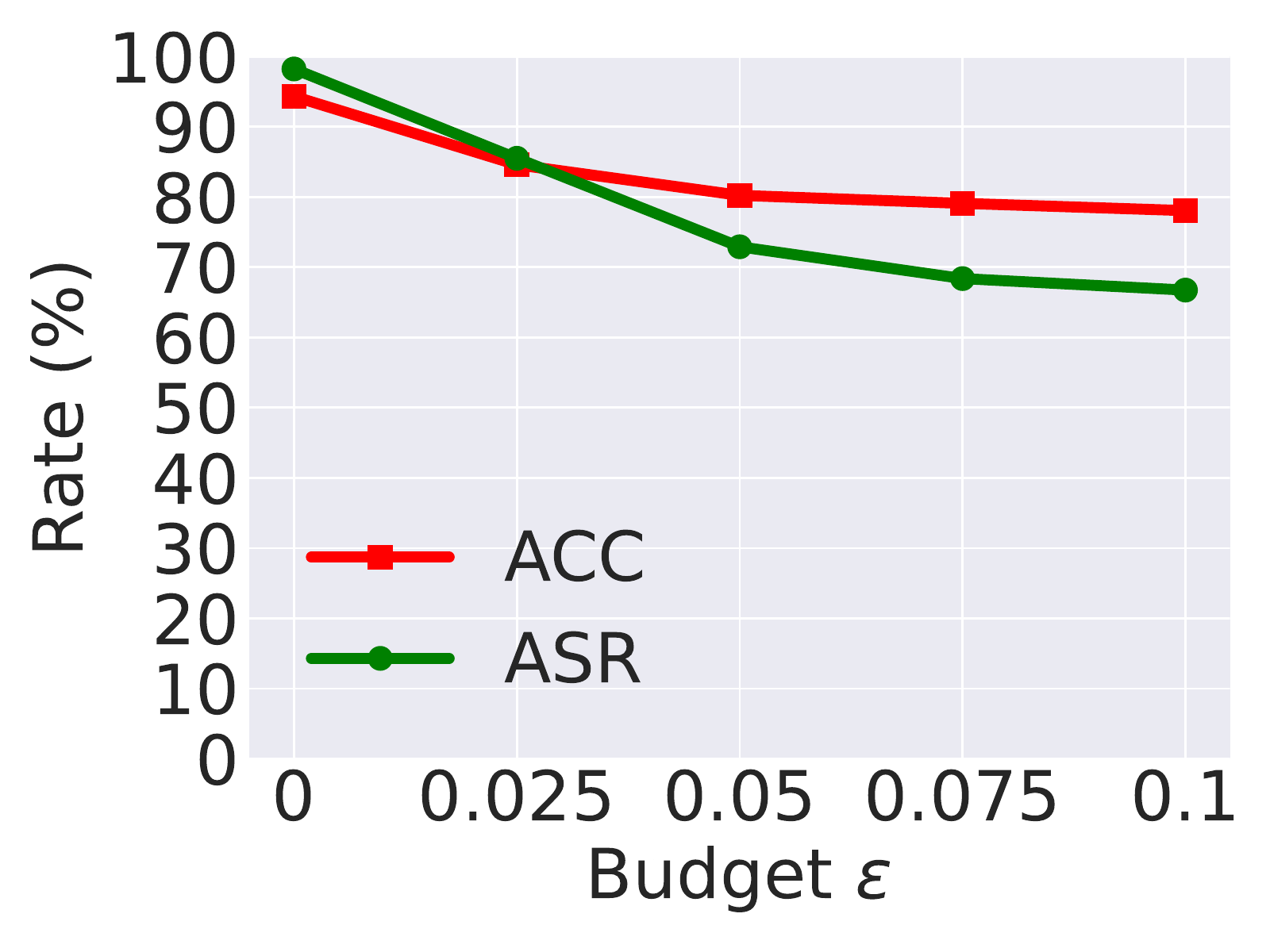}}
\subfigure[{WaNet, Spatial}]{
\label{fig:wanet_spatial}
\includegraphics[scale=0.22]{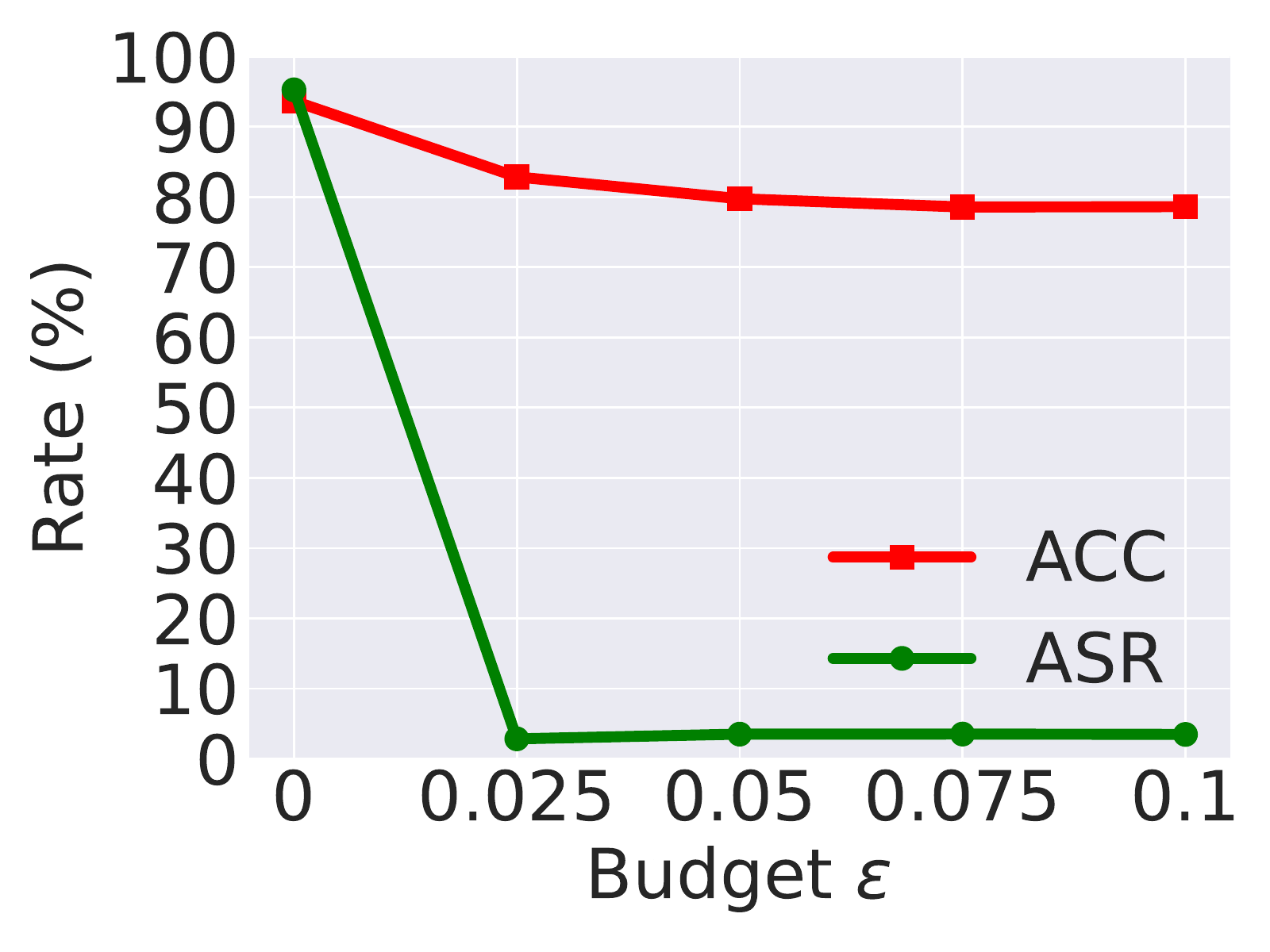}}

\subfigure[BadNets, Perceptual]{
\label{fig:badnets_percept}
\includegraphics[scale=0.22]{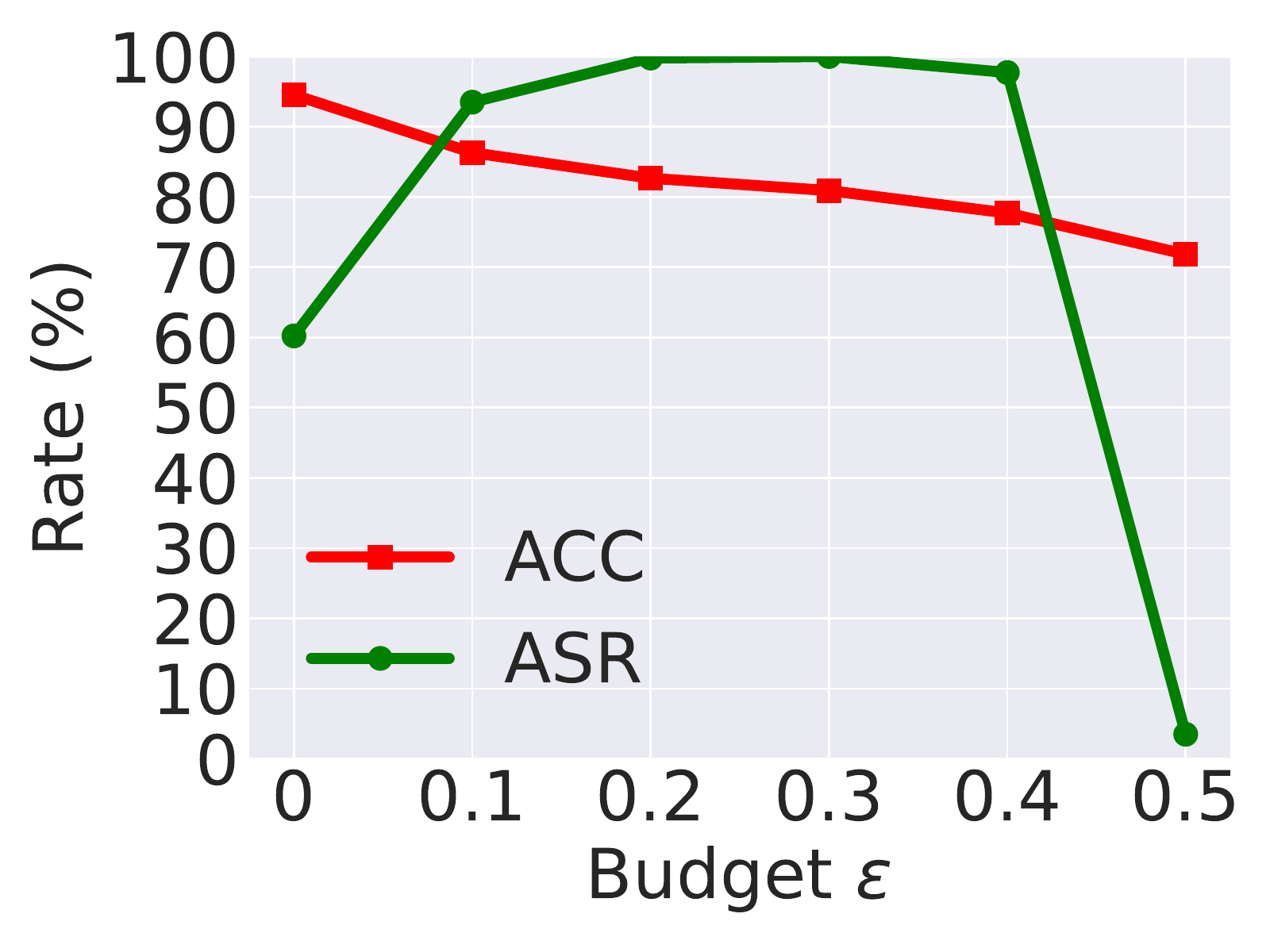}}
\subfigure[{LC, Perceptual}]{
\label{fig:lc_percept}
\includegraphics[scale=0.22]{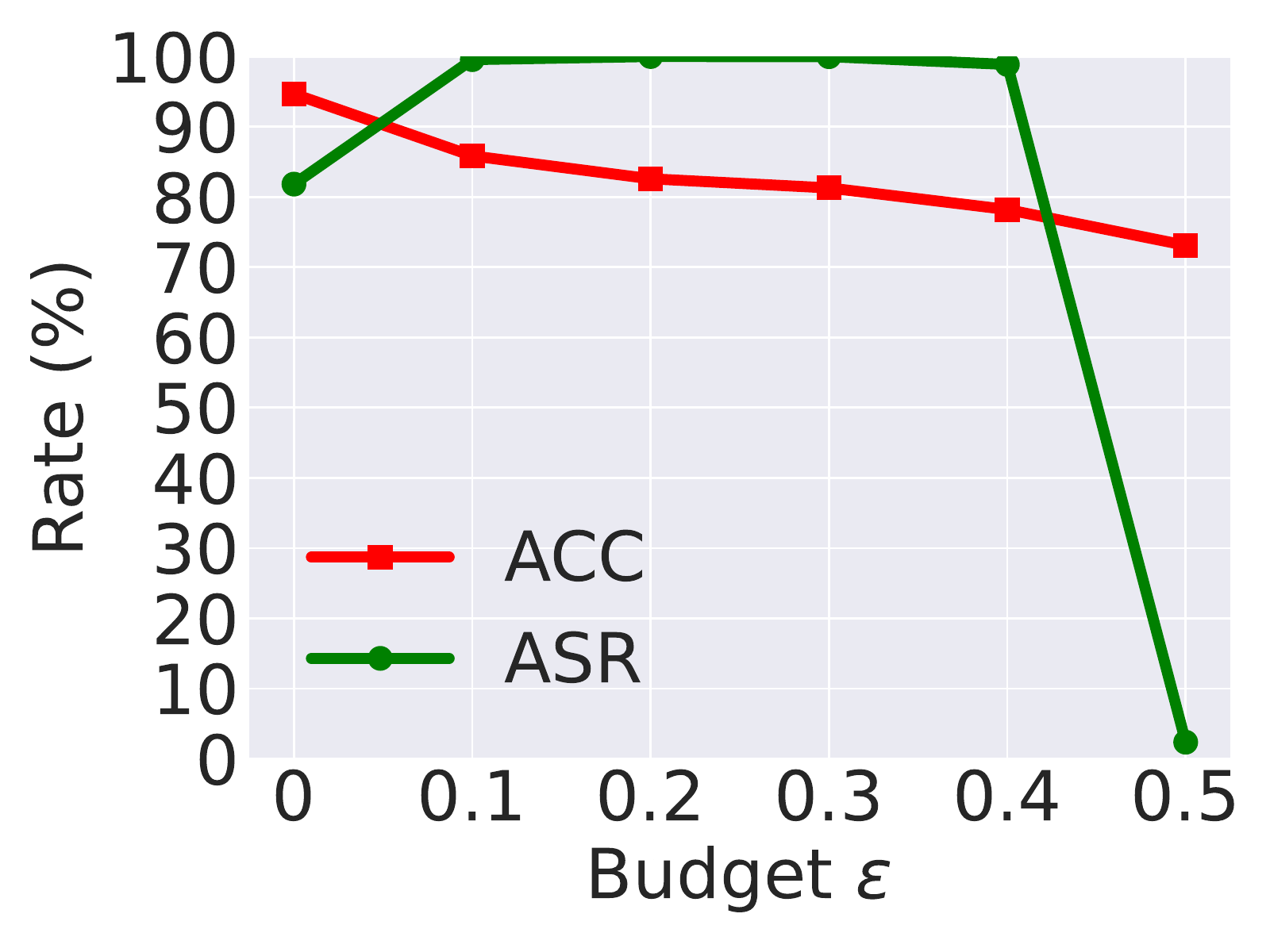}}
\subfigure[Blended, Perceptual]{
\label{fig:blended_percept}
\includegraphics[scale=0.22]{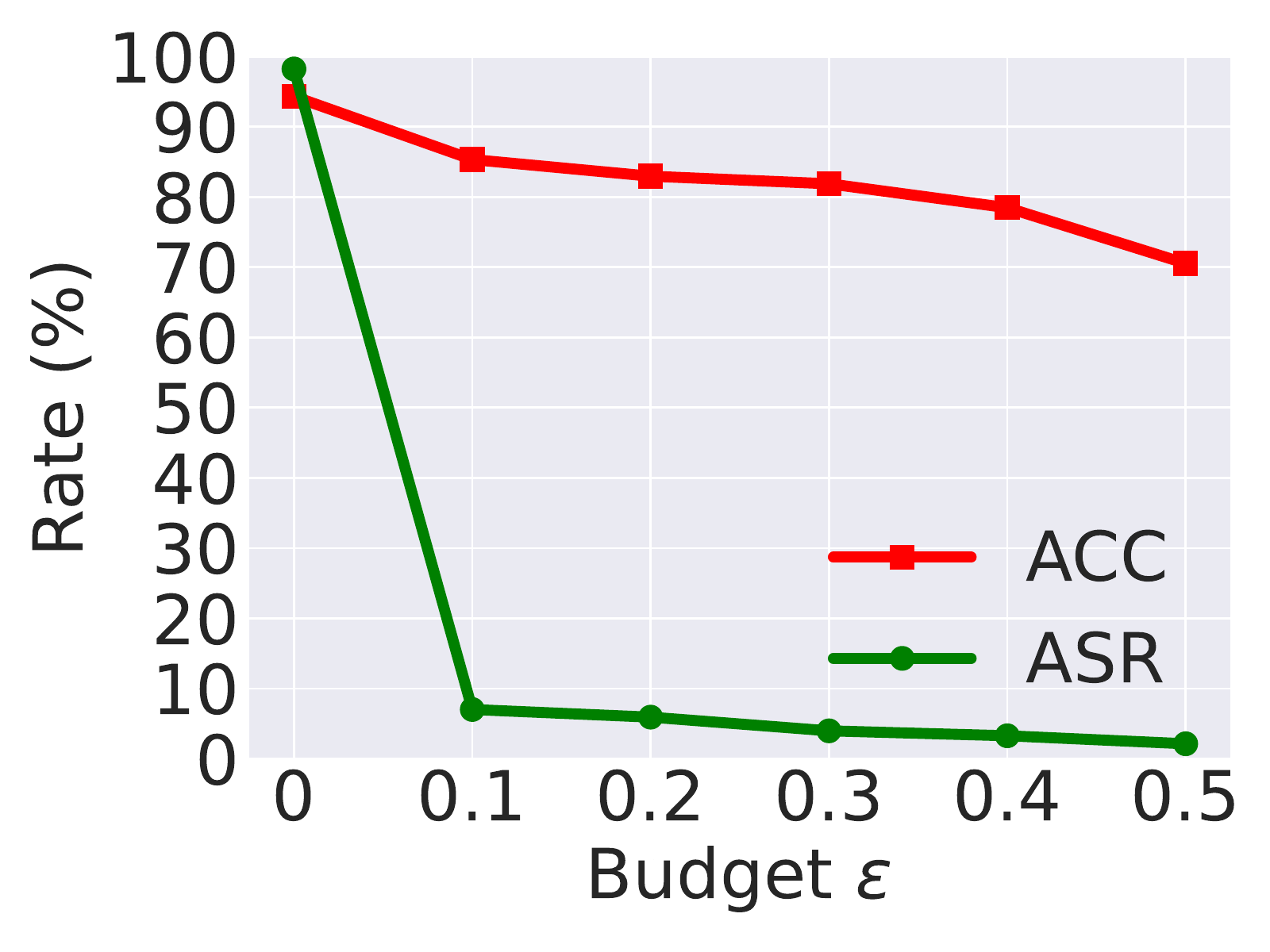}}
\subfigure[{WaNet, Perceptual}]{
\label{fig:wanet_percept}
\includegraphics[scale=0.22]{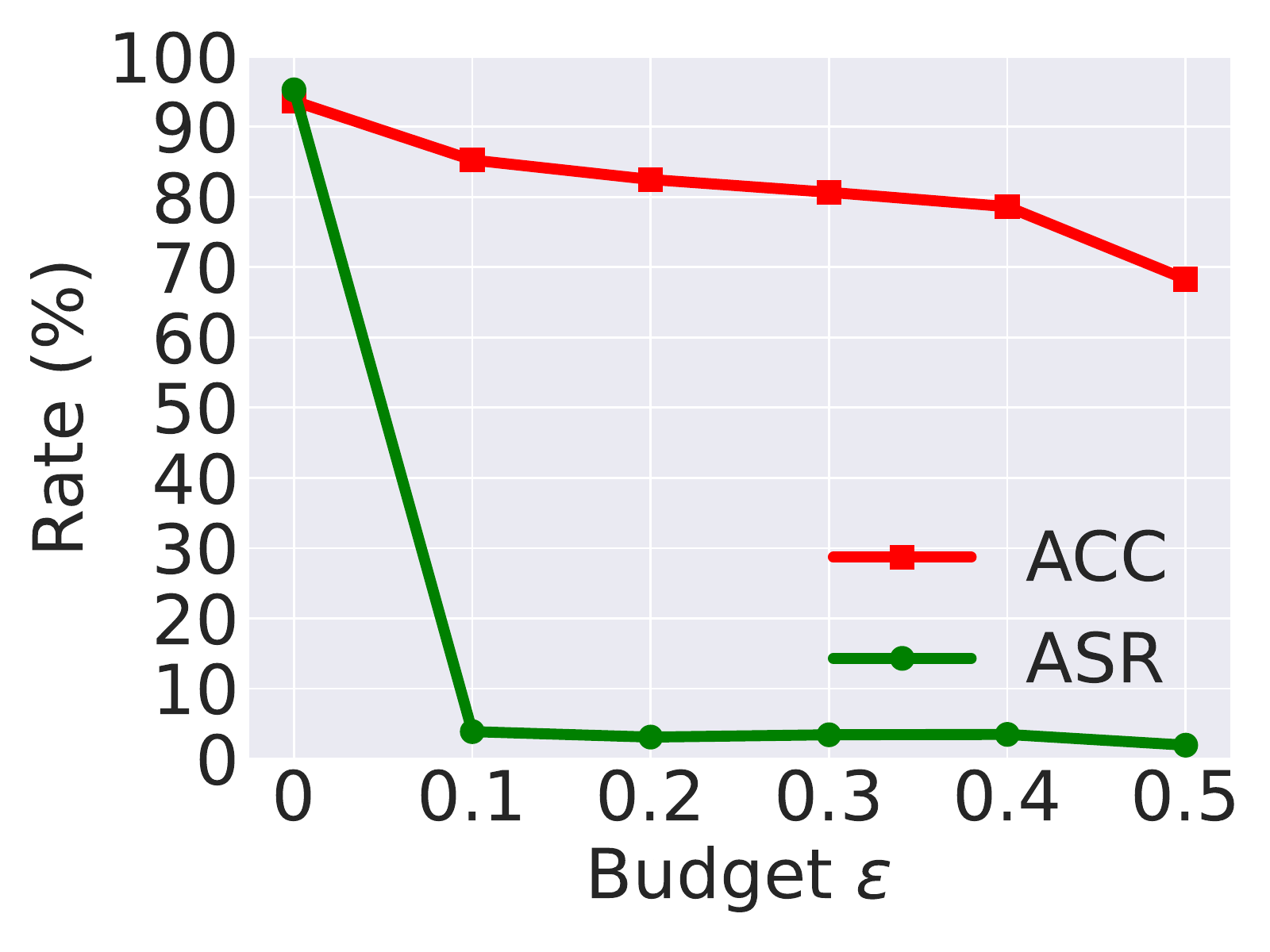}}

\caption{Evaluation of different backdoor attacks with different ATs on CIFAR-10.}
\label{fig1}
\end{figure*}

\begin{figure*}[!htbp]
\centering
\subfigure[BadNets, $L_\infty $]{
\label{fig:cifar100_badnets_linf}
\includegraphics[scale=0.22]{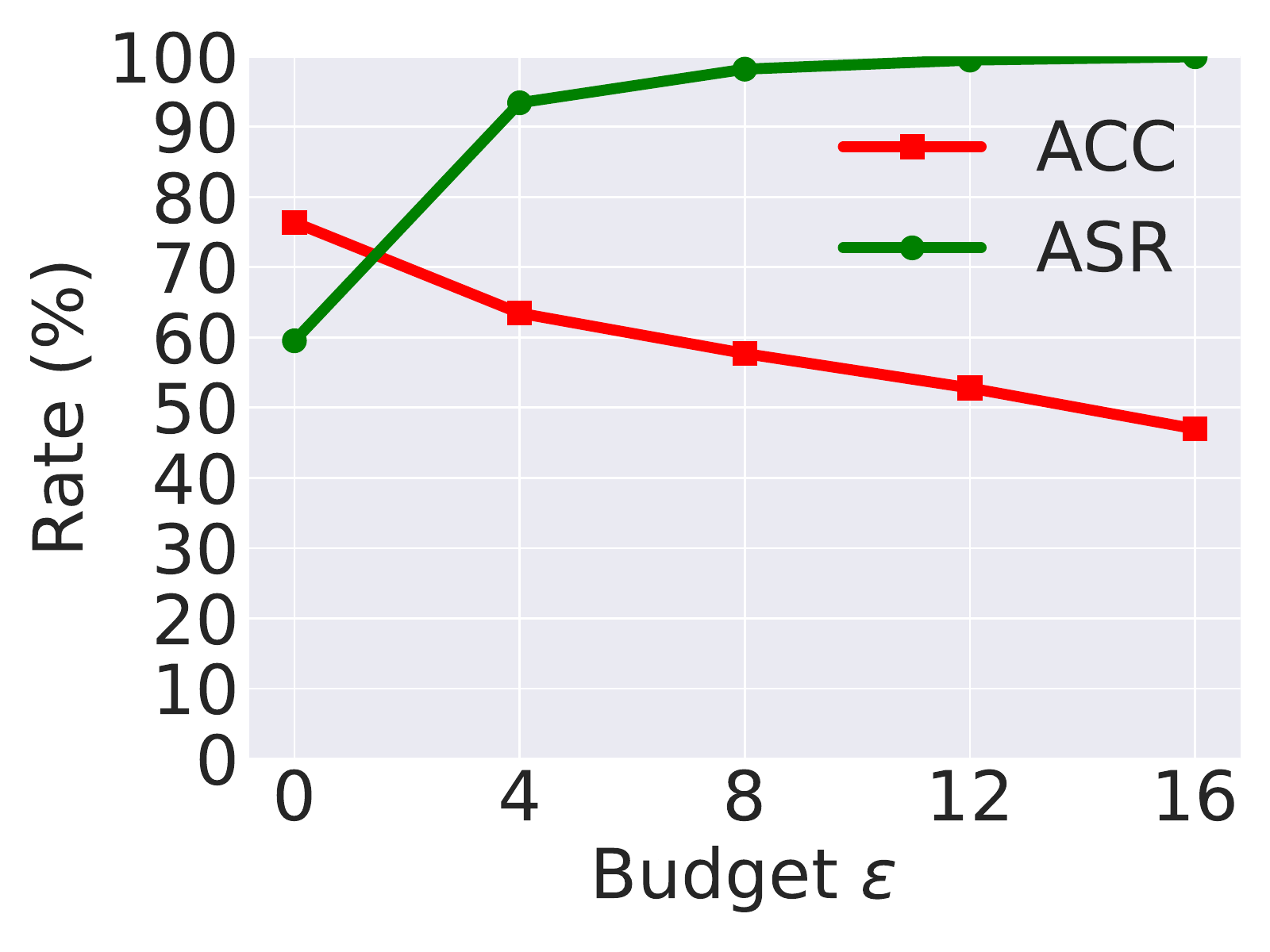}}
\subfigure[ {{LC, $L_\infty $}}]{
\label{fig:cifar100_lc_linf}
\includegraphics[scale=0.22]{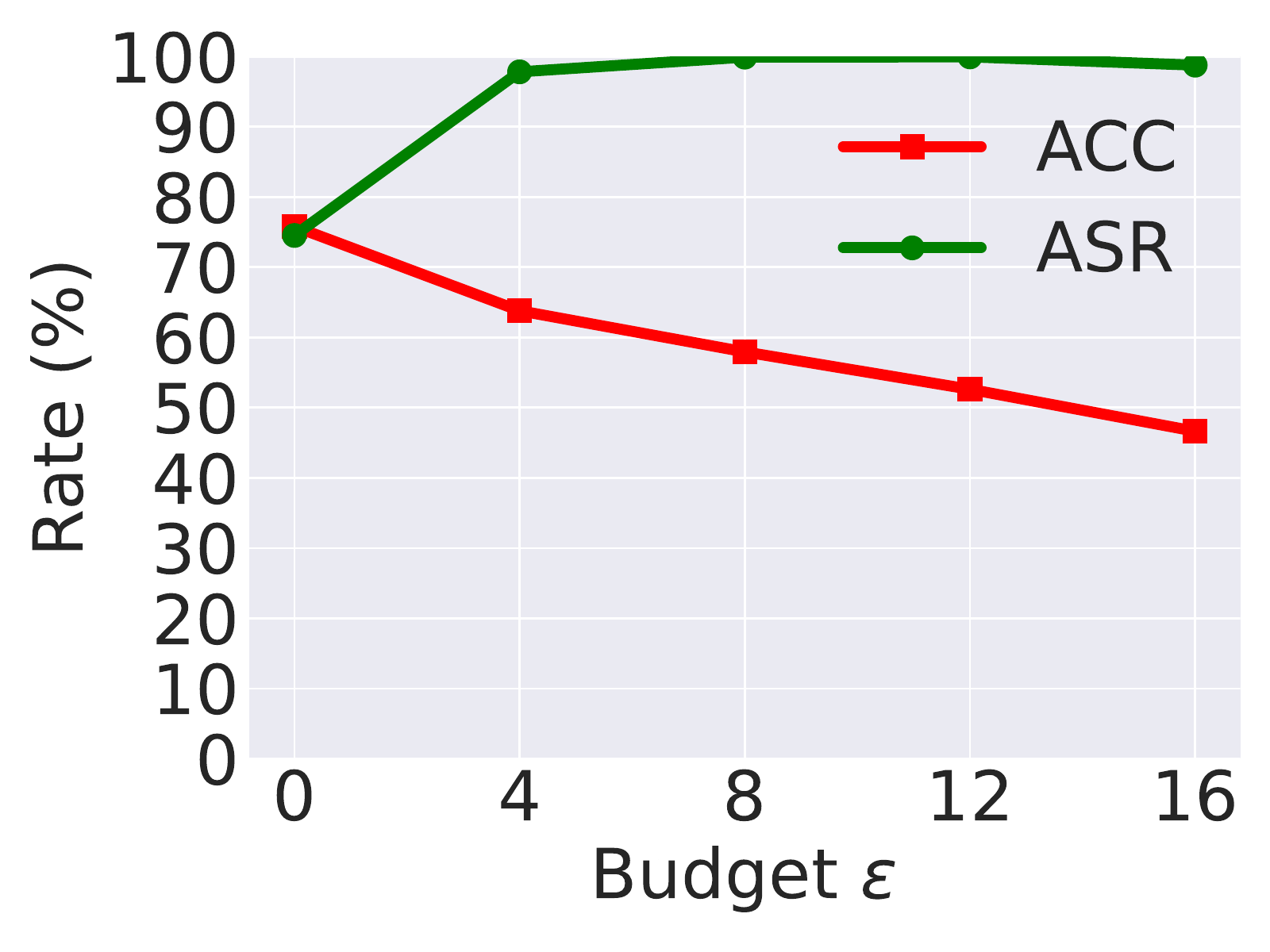}}
\subfigure[Blended, $L_\infty $]{
\label{fig:cifar100_blended_linf}
\includegraphics[scale=0.22]{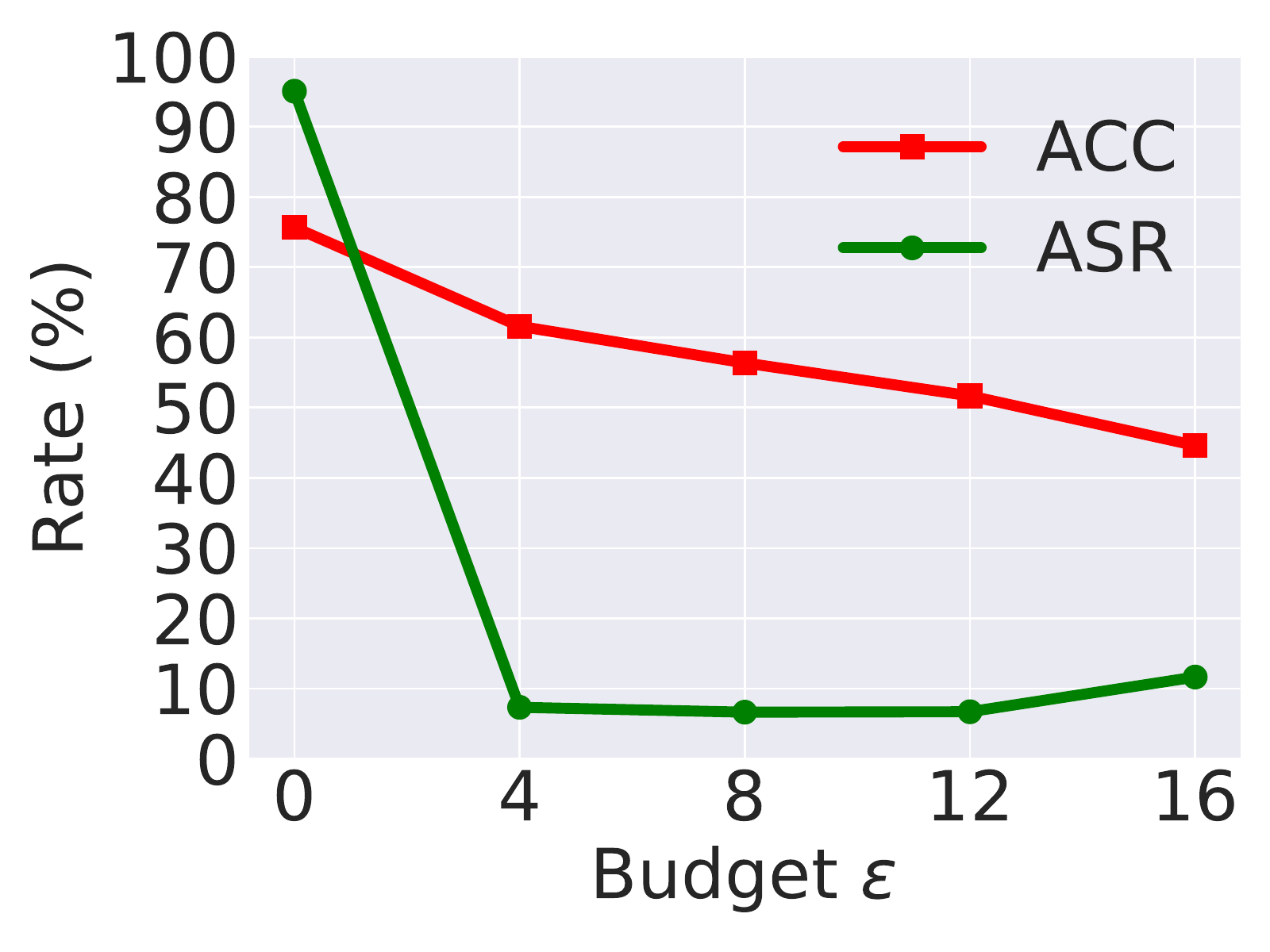}}
\subfigure[WaNet, $L_\infty $]{
\label{fig:cifar100_wanet_linf}
\includegraphics[scale=0.22]{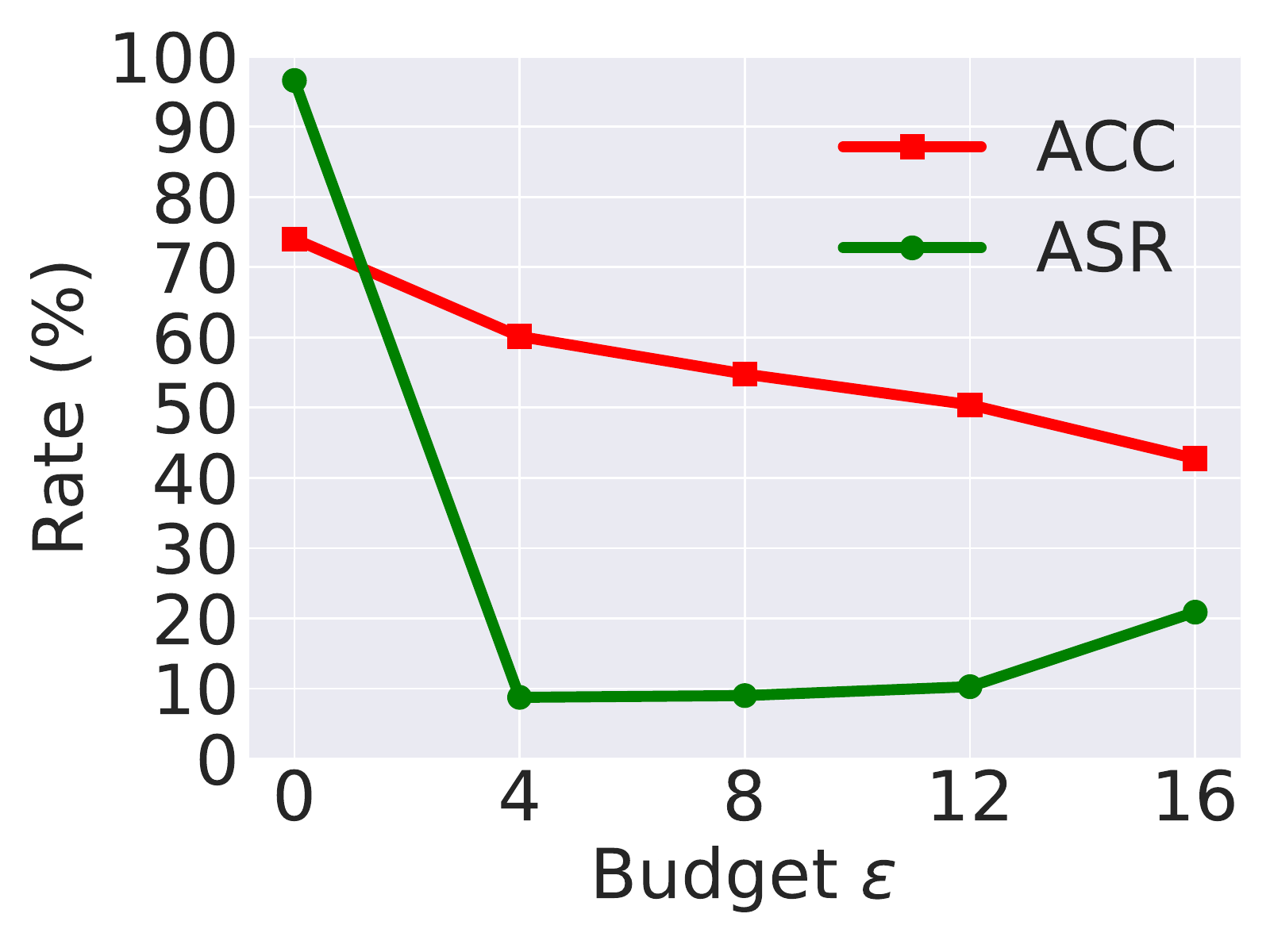}}

\subfigure[BadNets, $L_2 $]{
\label{fig:cifar100_badnets_l2}
\includegraphics[scale=0.22]{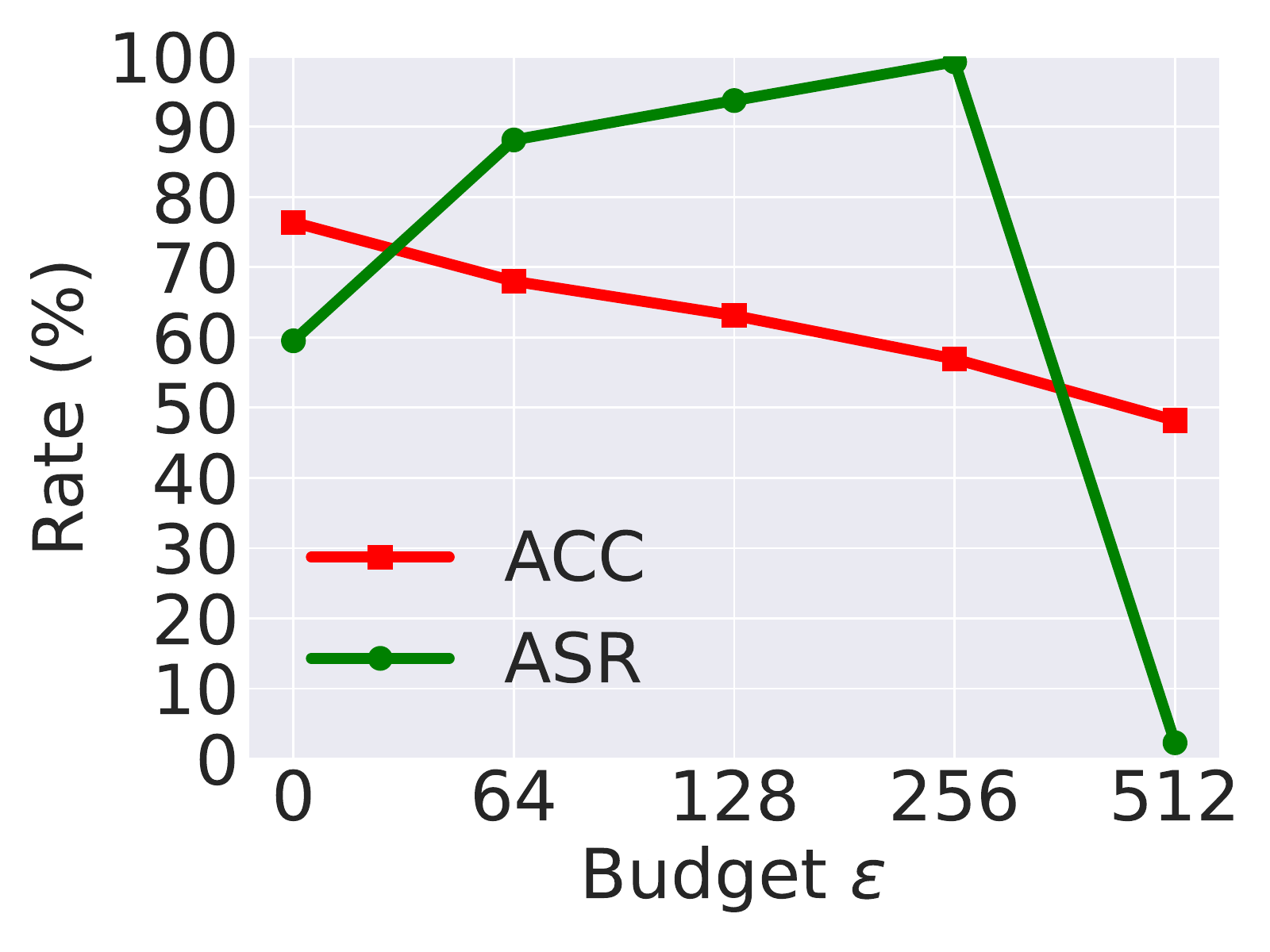}}
\subfigure[{{LC, $L_2$}}]{
\label{fig:cifar100_lc_l2}
\includegraphics[scale=0.22]{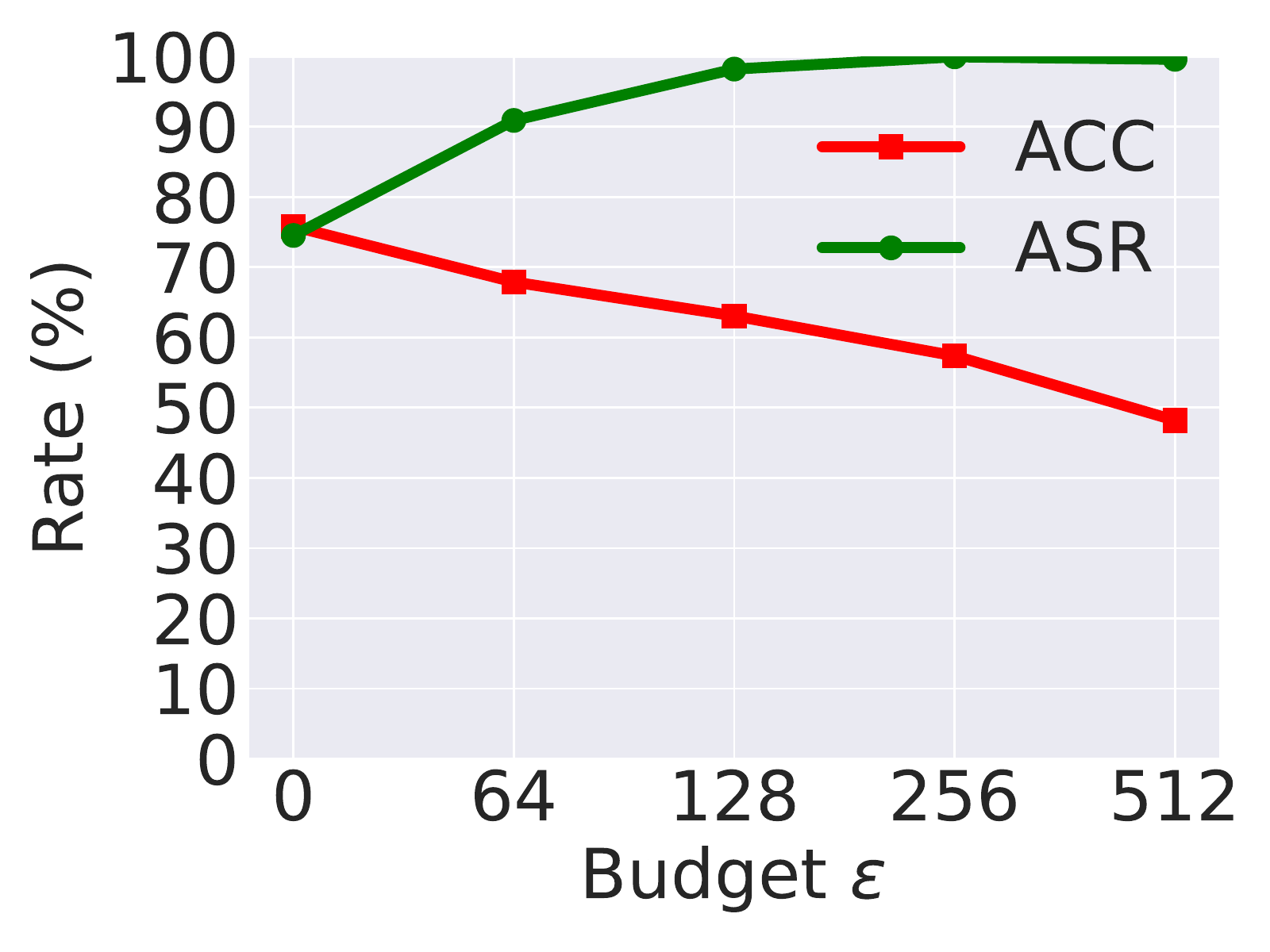}}
\subfigure[Blended, $L_2 $ ]{
\label{fig:cifar100_blended_l2}
\includegraphics[scale=0.22]{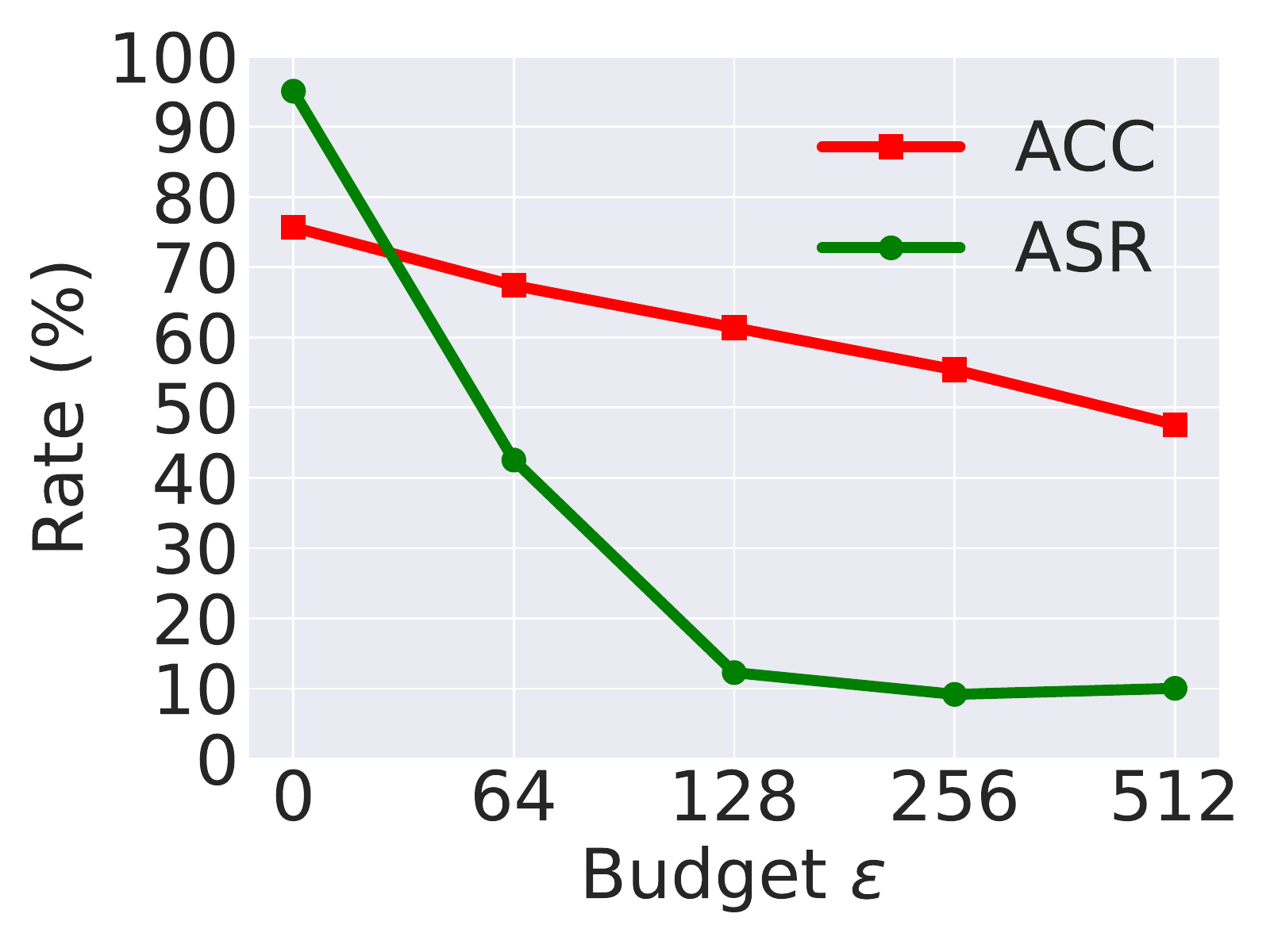}}
\subfigure[WaNet, $L_2$]{
\label{fig:cifar100_wanet_l2}
\includegraphics[scale=0.22]{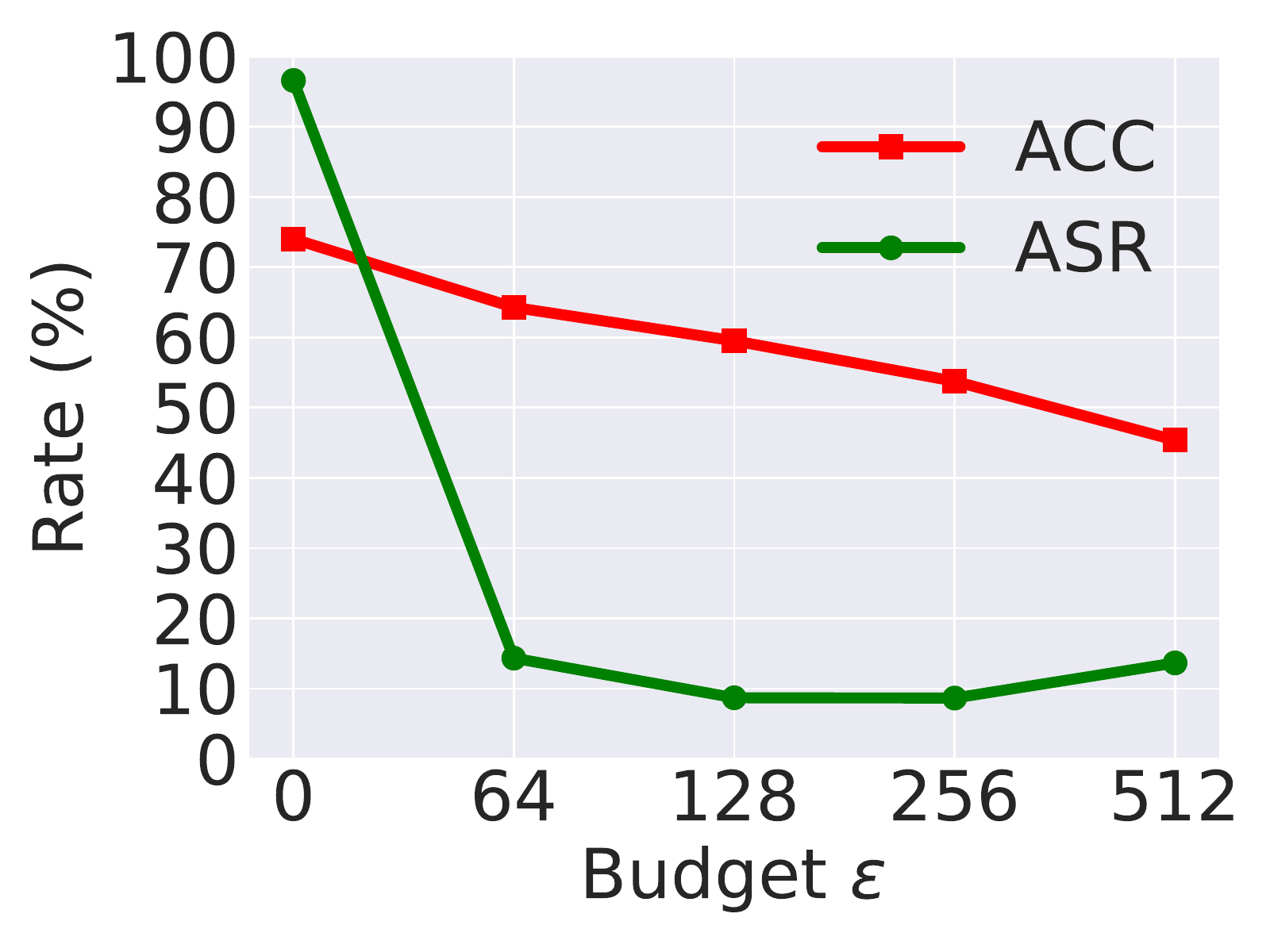}}

\subfigure[BadNets, Spatial]{
\label{fig:cifar100_badnets_spatial}
\includegraphics[scale=0.22]{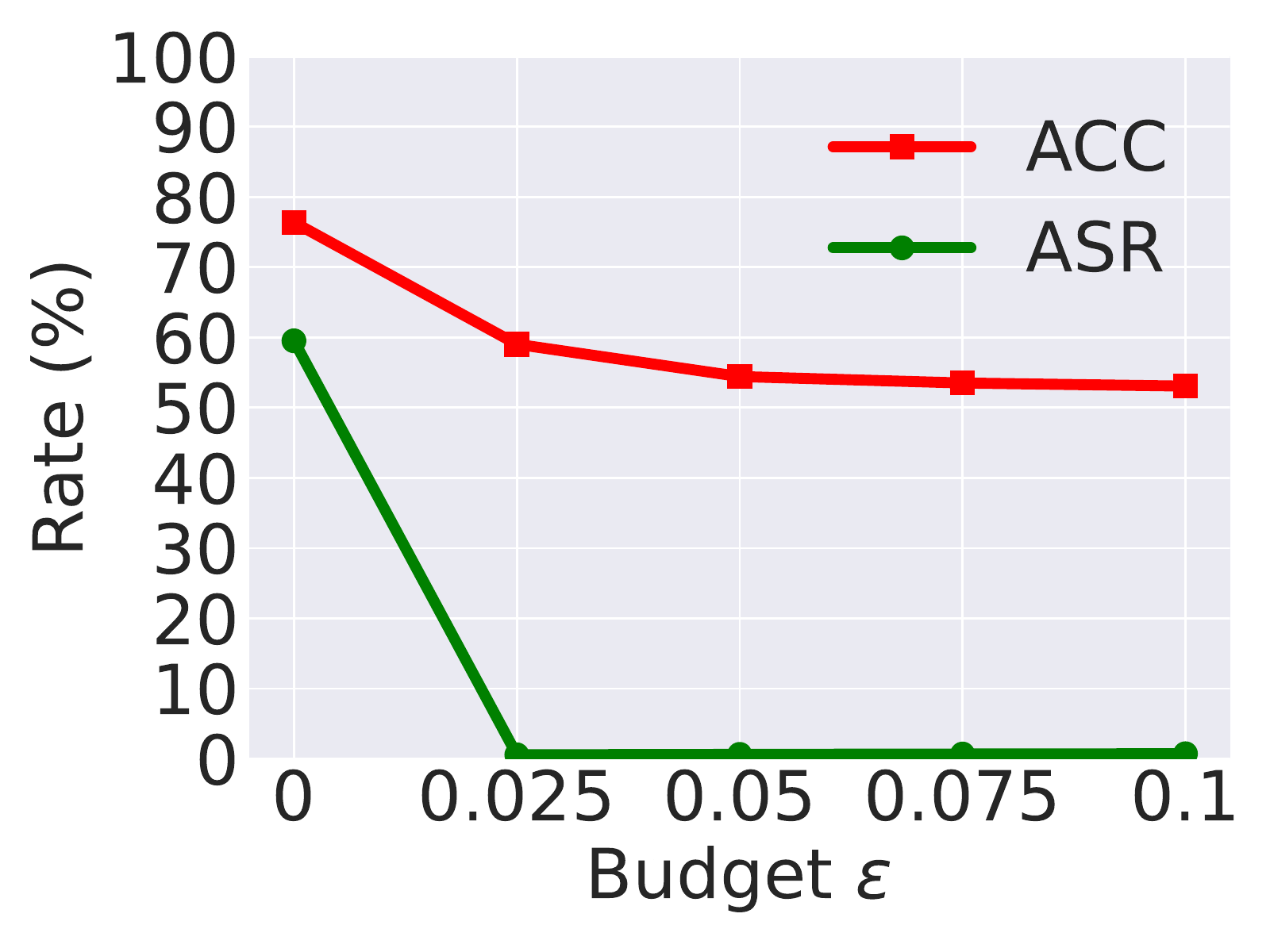}}
\subfigure[{LC, Spatial}]{
\label{fig:cifar100_lc_spatial}
\includegraphics[scale=0.22]{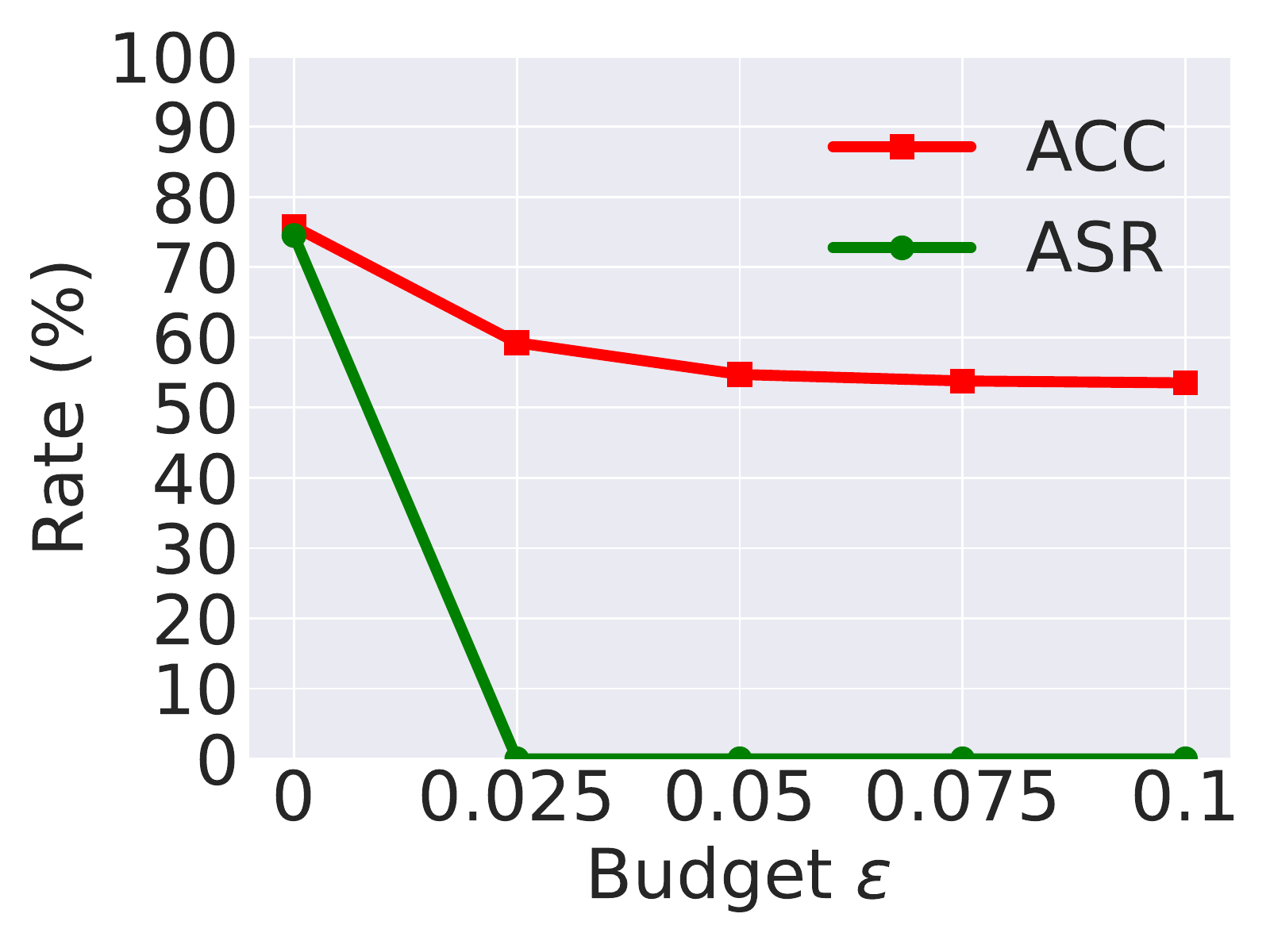}}
\subfigure[Blended, Spatial]{
\label{fig:cifar100_blended_spatial}
\includegraphics[scale=0.22]{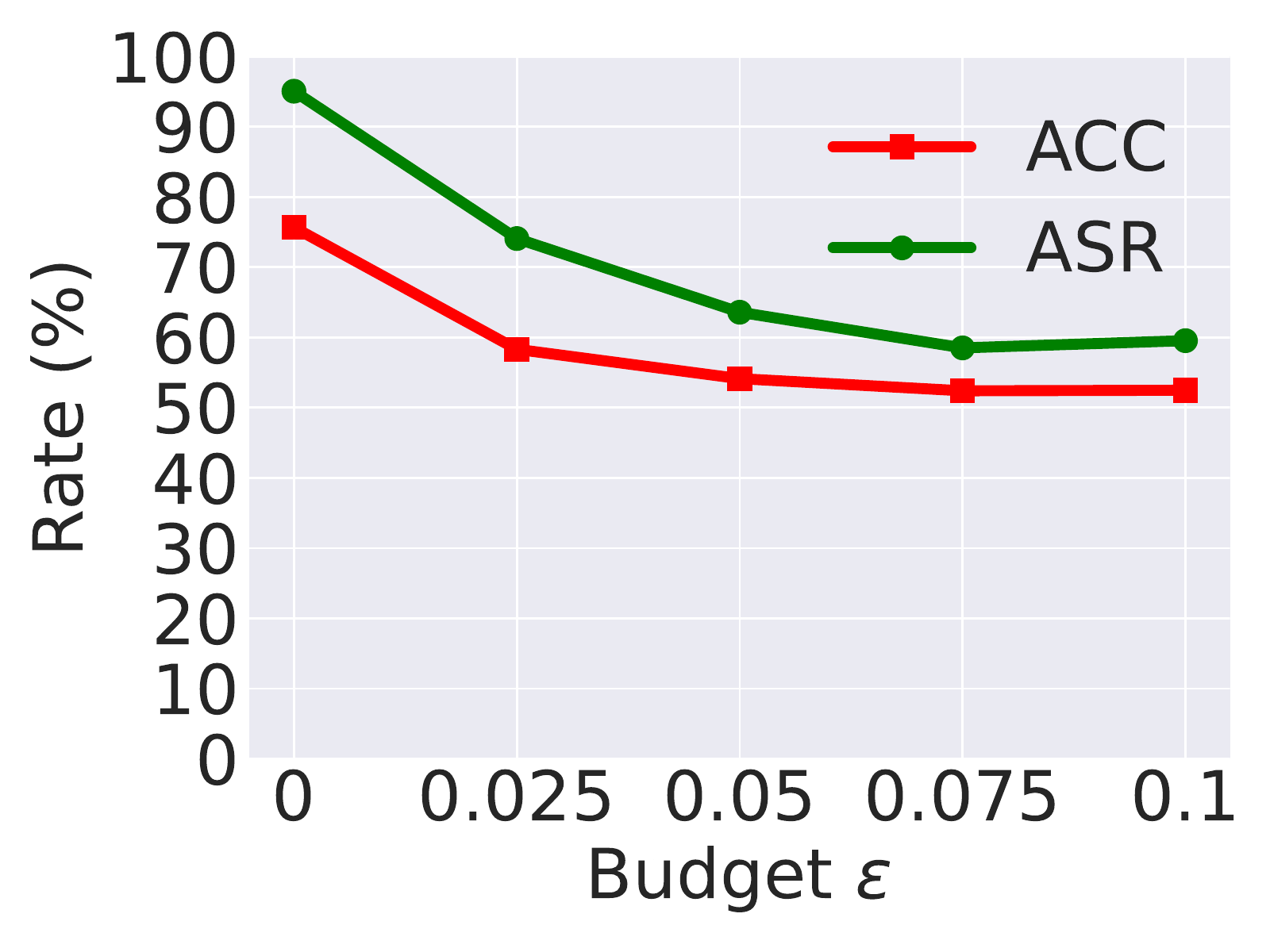}}
\subfigure[{WaNet, Spatial}]{
\label{fig:cifar100_wanet_spatial}
\includegraphics[scale=0.22]{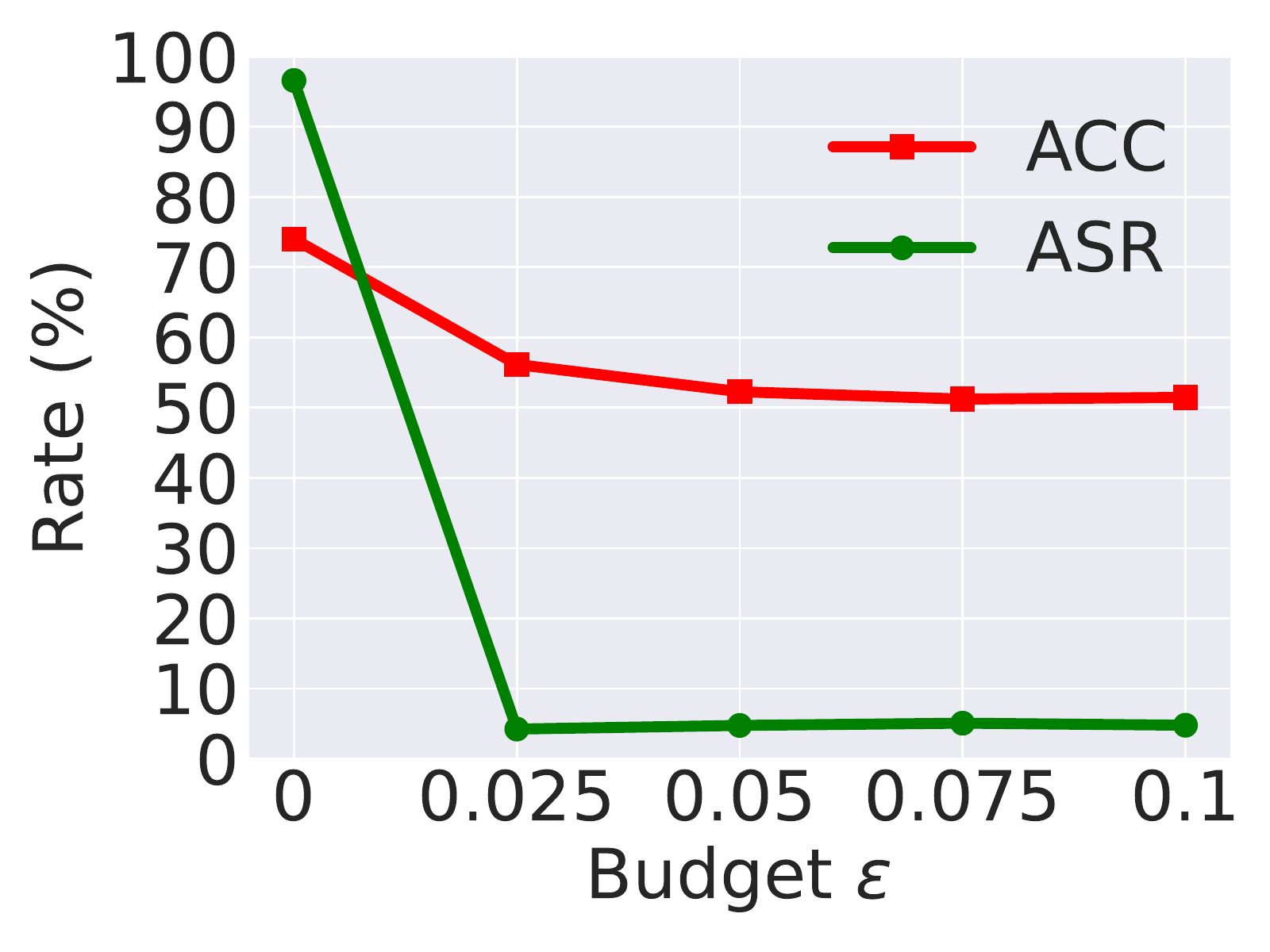}}

\subfigure[{ BadNets, Perceptual}]{
\label{fig:cifar100_badnets_percept}
\includegraphics[scale=0.22]{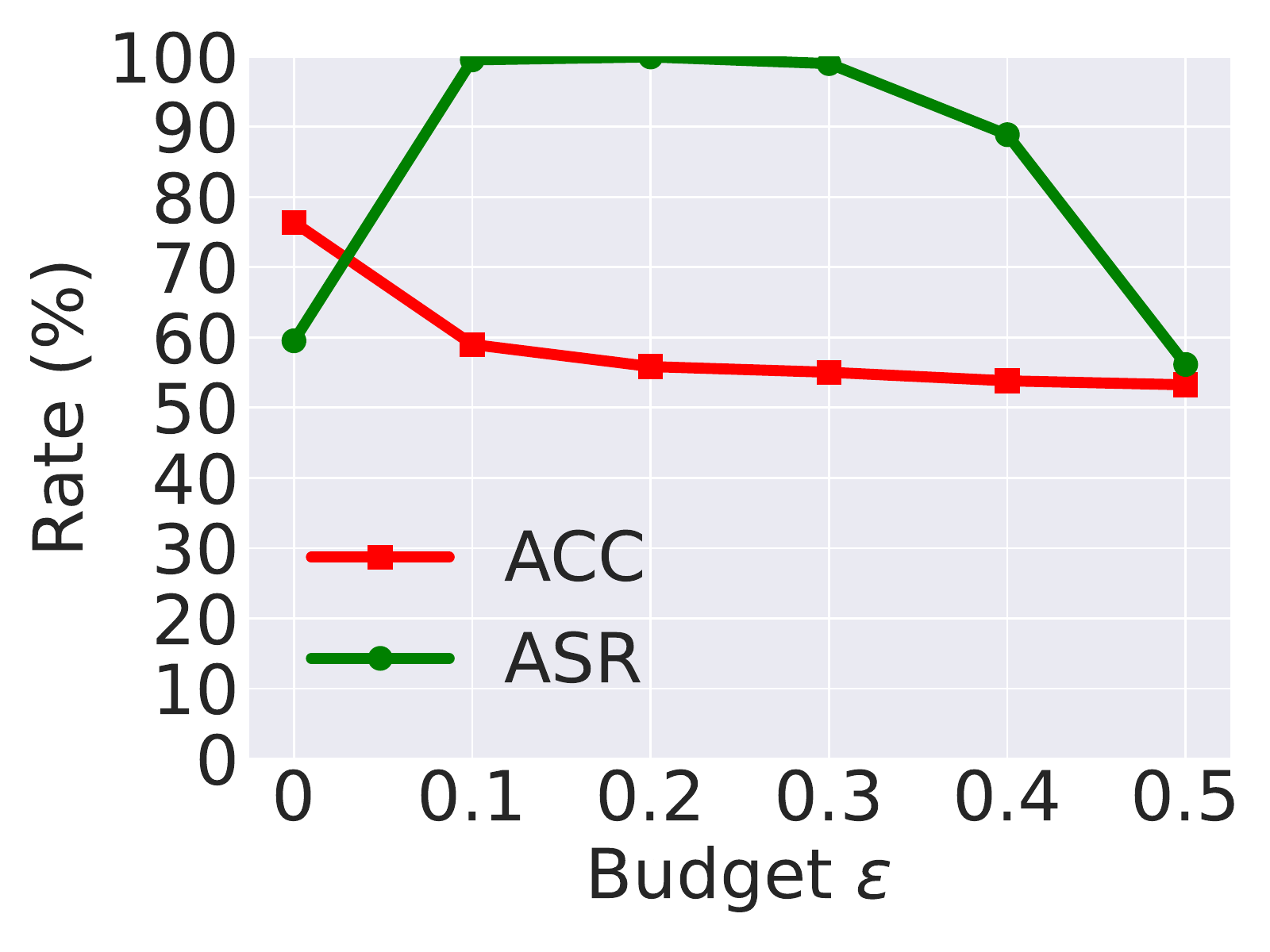}}
\subfigure[{LC, Perceptual}]{
\label{fig:cifar100_lc_percept}
\includegraphics[scale=0.22]{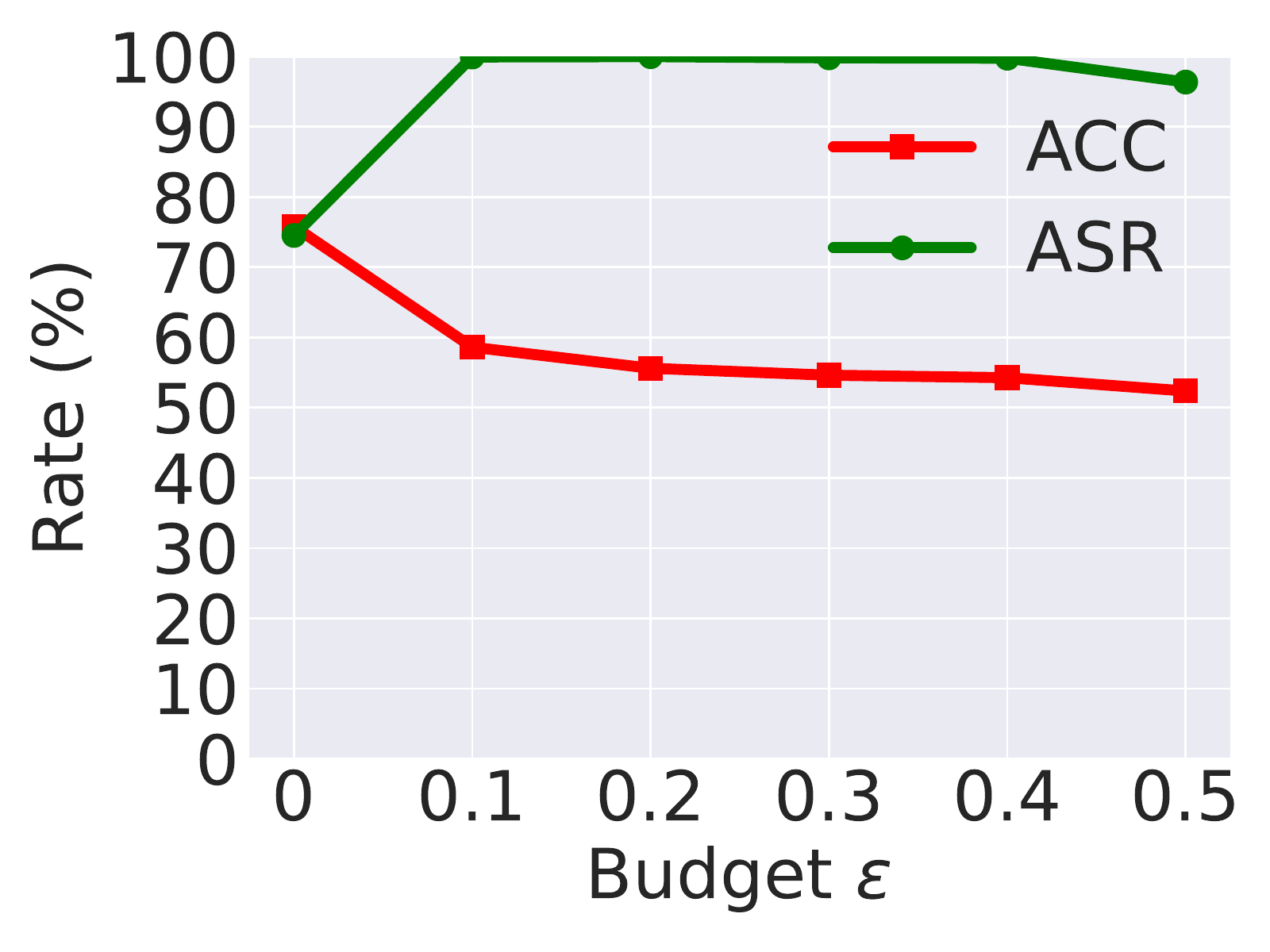}}
\subfigure[{Blended, Perceptual}]{
\label{fig:cifar100_blended_percept}
\includegraphics[scale=0.22]{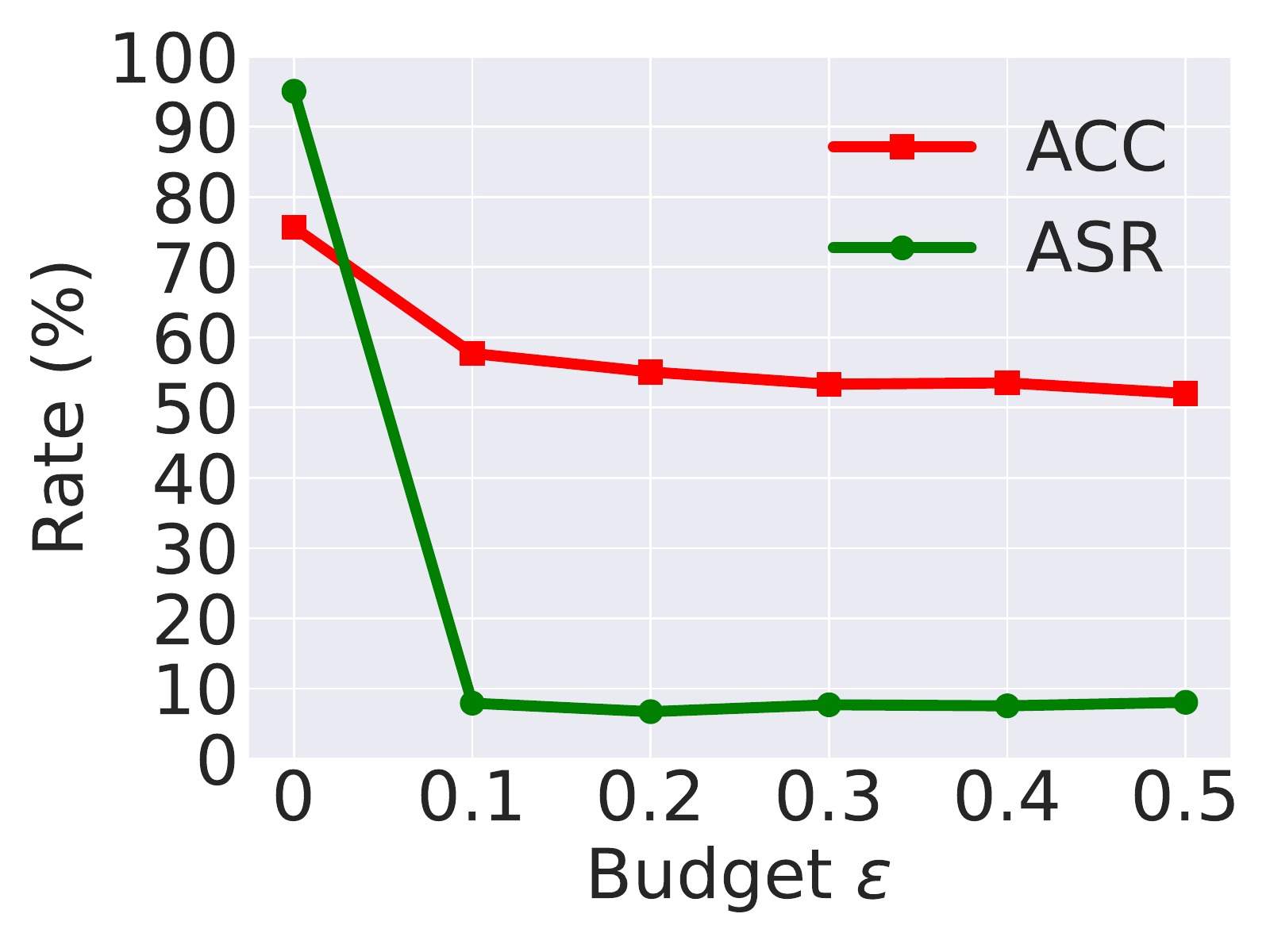}}
\subfigure[{WaNet, Perceptual}]{
\label{fig:cifar100_wanet_percept}
\includegraphics[scale=0.22]{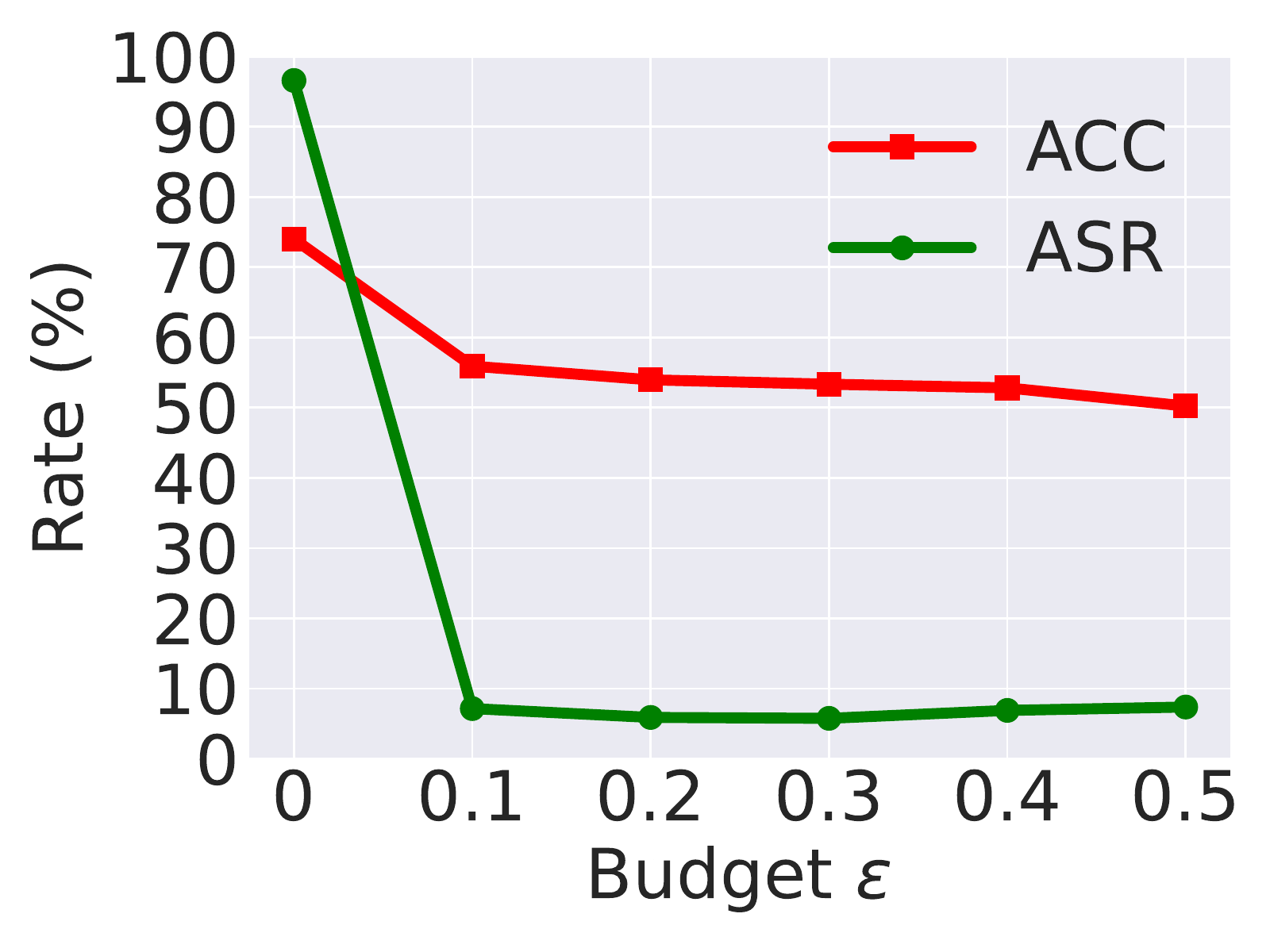}}

\caption{Evaluation of different backdoor attacks with different ATs on CIFAR-100. }
\label{fig2:cifar100}
\end{figure*}

We first adversarially trained a model on poisoned training data with different threat models across different backdoor attacks, and then illustrate how ACC and ASR of the trained model change with respect to different perturbation budgets in Figure \ref{fig1} (on CIFAR-10) and Figure \ref{fig2:cifar100} (on CIFAR-100). Interestingly, we have following observations:
\begin{itemize}[leftmargin=*]
\item \textit{Results Vary with Different Perturbation Budgets.} In Figures \ref{fig:badnets_linf}-\ref{fig:lc_linf}, we observe that, when $\epsilon \leq 12/255$, the ASR increases with larger perturbation budgets in commonly-used $L_\infty$ threat models, which is consistent to the phenomenon that AT indeed strengthened the backdoor robustness in Weng \textit{et al.} \citep{weng2020trade}. However, if the perturbation budget continues to increase ($\epsilon > 16/255$), the ASR starts to decrease, which means AT could still could mitigate the backdoor vulnerability as long as the perturbation budget is large enough. Therefore, the findings in Weng \textit{et al.} \citep{weng2020trade} are actually incomplete since they ignored the effects of perturbation budgets.

\item \textit{Threat Models Matter for Backdoor Defense.} In Figure \ref{fig:badnets_linf} on CIFAR-10, even though the ACC drops to $\thicksim 80\%$ in $L_{\infty}$ AT with $\epsilon=12/255$, the ASR still achieves $100\%$. Only when we enlarged the perturbation budget to $\epsilon=16/255$ with only $\thicksim 70\%$ of ACC, we obtained  satisfactory backdoor robustness (close $0\%$ of ASR). Instead, we could easily achieve $\thicksim 85\%$ of ACC and $\thicksim 0\%$ of ASR via spatial AT (budget $\epsilon=0.025$) in Figure \ref{fig:badnets_spatial}. Further, while $L_p$ AT achieved unsatisfactory performance under BadNets and LC, we find spatial AT easily mitigates these backdoors. Unfortunately, spatial AT does not work well under Blended Attack, different from others. This indicates that threat models matter in backdoor defense. We conjecture that spatial adversarial transformation can easily distort the trigger pattern (see Figure \ref{fig:adv_back}), making the trained model keep the prediction unchanged in the presence of the trigger pattern. As a result, we could achieve satisfactory performance if we select a suitable threat model. Similar phenomenon can be found on CIFAR-100 in Figure \ref{fig2:cifar100}. In addition, WaNet \citep{nguyen2021wanet}, a SOTA backdoor attack, is fragile and easily mitigated by $L_p$
/ spatial / perceptual adversarial perturbations, which reminds researchers of not only considering the stealthiness of backdoor attacks, but also their durability and persistence against backdoor defenses.

\end{itemize}

\begin{table}[h!]
\caption{Comparison with random perturbations. }
\begin{center}
\scalebox{0.9}{
\begin{tabular}{c|cc}
\toprule
 & ACC & ASR  \\ 
\midrule
BadNets ($3\times 3$) & 94.82 & 100.00 \\
Spatial AT ($\epsilon=0.025$) & 84.21 & 3.03  \\
Random perturbation ($\sigma=0.05$) &  89.73 &77.17 \\
Random perturbation ($\sigma=0.1$) &  83.68 &44.98 \\
Random perturbation ($\sigma=0.15$) &  80.49 &9.89 \\
Random perturbation ($\sigma=0.2$) &  63.92 &14.23 \\
\midrule
Blended (Hello-Kitty) & 94.50 & 98.47 \\
$L_\infty$ AT ($\epsilon=4/255$) & 87.17 & 4.56  \\
Random perturbation ($\sigma=0.05$) &  91.70 &88.82 \\
Random perturbation ($\sigma=0.1$) &  86.92 &56.86 \\
Random perturbation ($\sigma=0.15$) &  83.28 &23.38 \\
Random perturbation ($\sigma=0.2$) &  79.38 &12.53 \\
\bottomrule
\end{tabular}
}
\end{center}
\label{tab:constrast}
\end{table}

{\bf Comparing to Random Perturbations}. From the findings above, AT indeed provides robustness against backdoor attacks at the cost of extra forward and backward propagation to calculate the adversarial perturbations, which is time-consuming. Naturally, if we could apply random perturbations to mitigate the backdoor vulnerability, the overhead from random perturbations is almost neglected. Here we explore whether random perturbations to input could defend against backdoor attacks or not. Specifically, we trained models with randomly perturbed input with varying budget and compared them with  adversarially trained models. During training, for Blended Attacks, we added the zero-mean Gaussian noise to the poisoned training set and compared the obtained model with the $L_\infty$ adversarially trained one. For BadNets, we used random spatial transformations and compared the obtained model with the adversarially trained one. As shown in Table \ref{tab:constrast}, for BadNets, we focused on  spatial AT and the randomly perturbed model ($\sigma=0.1$), as both models had similar clean accuracies ($\thicksim 84\%$). At this point, spatial AT has successfully mitigate backdoor attacks (the ASR is below to $5\%$) while the randomly perturbed model does not (the ASR is $44.98\%$ ). Similar observations also hold for Blended Attack. Then, we draw the conclusion that adversarial perturbations are superior to random perturbations in the terms of backdoor robustness.

\section{Composite Adversarial Training}
\label{section-cat}

We have demonstrated that AT with varying threat models performs differently under varying backdoor attacks. For example, spatial AT effectively defends against patch-based attacks, while $L_\infty$ defends against whole-image attacks. However, in real scenarios, we have no knowledge about the trigger pattern. Therefore, we propose \textit{Composite Adversarial Training} (CAT) which integrates two effective adversarial perturbations: $L_\infty$ adversarial attacks and spatial transformation attacks, the former for mitigating the global-perturbation attacks and the latter for local-patch attacks. 

{\bf Experimental Settings.} We used $\epsilon=2/255$ for $L_\infty$ AT and $\epsilon=0.025$ for spatial AT considering the trade-off between natural accuracy and robustness. We evaluated our method on CIFAR-10 and CIFAR-100 against the four SOTA backdoor attacks. For CIFAR-10, the attack settings were the same as before except that we adopted the dirty label setting for BadNets and the poison rate was increased to $1\%$ for BadNets and LC. Settings for CIFAR-100 were similar and we leave the details in Appendix \ref{app:cifar100}.


{\bf Baseline Methods.}
We compared CAT with a series of backdoor defense methods: Fine Pruning (FP) \citep{liu2018fine}, Neural Attention Distillation (NAD) \citep{li2021neural}, DPSGD \citep{hong2020effectiveness} and ABL \citep{li2021anti}. We grid-searched the pruning ratio for FP, from 5\% to 95\% with step 5\%, and chose the result whose clean accuracy is closest to ours for a fair comparison. For NAD, we followed the original settings but set the initial learning rate to 0.01 for more stable results. For DPSGD, we replaced batch normalization with group normalization to obey the rule of differential privacy and set the noise level $\sigma$ to 0.1. For ABL, we adopted the same settings in its paper except for the loss threshold $\gamma$, which we set to 0 for a better detection rate. 


\begin{table*}[!htbp]
  \caption{Comparison with other backdoor defense methods on CIFAR-10. The lowest ASR is indicated in boldface and the second-lowest ASR is indicated with an underline.  }
  \vskip 0.15in
  \centering
  \scalebox{0.9}{
  \begin{tabular}{@{}l|cc|cc|cc|cc|cc|cc@{}}
    \toprule
    &\multicolumn{2}{c|}{No defense } &\multicolumn{2}{c|}{FP }&\multicolumn{2}{c|}{NAD} &\multicolumn{2}{c|}{DPSGD } &\multicolumn{2}{c|}{ABL } &\multicolumn{2}{c}{CAT }\\
    \cmidrule(r){2-3}\cmidrule(r){4-5}\cmidrule(r){6-7}\cmidrule(r){8-9}\cmidrule(r){10-11} \cmidrule(r){12-13}
    Attack & ACC & ASR & ACC & ASR & ACC & ASR & ACC & ASR & ACC & ASR & ACC & ASR\\
    \midrule
    BadNets($3\times 3$) & 94.68 & 100.00 & 94.14 & 99.92 & 89.35 & \underline{10.19} & 71.40 & 99.90 & 79.58 & 94.28 & 81.28 & 
    {\bf 2.60}  \\
 BadNets($2\times 2$)  & 94.35 & 100.00 & 94.29 & 93.81 & 89.99 & \underline{7.68} &71.70 &99.69 & 82.23 & 98.83 & 81.51 & {\bf 1.83}  \\
 BadNets(Random) & 94.68 & 98.54 & 94.34 & 96.25 & 89.39 & 18.62 &71.14 &\underline{4.13} & 80.05 & 88.01 & 82.01 & {\bf 1.40}  \\
 LC      & 94.47 & 88.64 & 85.46 & {\bf 0.00} & 79.81 & \underline{0.05} & 71.24 & 99.40 & 83.44 & {\bf 0.00} & 82.20 & 0.18   \\
 Blended(Hello-Kitty)     & 94.50 & 98.47 & 94.21 & 71.21 & 89.45 & \underline{10.02} &70.41 &67.93 & 78.57 & 47.70 & 82.11 & {\bf 3.74} \\
Blended(Random)      & 94.59 & 100.00 & 94.24 & 98.81 & 94.36 & 5.54 &70.06 &99.92 & 79.16 & {\bf 0.70} & 80.71 & \underline{5.48} \\

WaNet      & 93.63 & 95.04 & 88.53 & 92.93 & 84.87 & \underline{2.28} & 69.22 &57.60 & 80.04 & 98.59 & 81.37 & {\bf 1.82}   \\
    \midrule  
Average     & 94.41 & 97.24 & 92.17 & 79.00 & 88.20 & \underline{7.77}  &70.74 &75.51 & 80.48 & 61.16 & 81.60 & {\bf 2.44} \\
    \bottomrule
  \end{tabular}
  }
  \vskip -0.05in
  \label{tab:main_cifar10}
\end{table*}

\begin{table*}[!htbp]
  \caption{Comparison with other backdoor defense methods on CIFAR-100.}
  \vskip 0.15in
  \centering
  \scalebox{0.9}{
  \begin{tabular}{l|cc|cc|cc|cc|cc|cc}
    \toprule
    &\multicolumn{2}{c|}{No defense } &\multicolumn{2}{c|}{FP }&\multicolumn{2}{c|}{NAD} &\multicolumn{2}{c|}{DPSGD } &\multicolumn{2}{c|}{ABL } &\multicolumn{2}{c}{CAT }\\
    \cmidrule(r){2-3}\cmidrule(r){4-5}\cmidrule(r){6-7}\cmidrule(r){8-9}\cmidrule(r){10-11} \cmidrule(r){12-13}
    Attack & ACC & ASR & ACC & ASR & ACC & ASR & ACC & ASR & ACC & ASR & ACC & ASR\\
    \midrule
    BadNets($3\times 3$) & 76.65 & 99.72 & 62.82 & 48.91 & 63.97 & 20.06 &24.47 &99.87 & 64.73 &{\bf 0.00} & 55.91 & 
     \underline{1.61}  \\
BadNets($2\times 2$)    &  76.45 & 99.65 & 60.94 & 92.92 & 59.47 & 6.34 &24.05 &\underline{4.28} & 66.48 & 71.93 & 57.07 & {\bf 0.57}  \\
BadNets(Random)   & 76.29 & 96.45 & 58.33 & 93.63 & 64.02 & 66.89 &24.58 &\underline{3.66}  &  65.62 & 76.39 & 56.91 &{\bf 0.46}  \\
LC & 75.92 & 98.71  & 59.44 & {\bf 0.00} & 60.8 & 0.19 &23.84 &95.88 & 67.75 & 85.95 & 56.71 & \underline{0.04} \\
Blended(Hello-Kitty)      & 75.84 & 94.45  & 59.76 & 14.25 & 57.86 & {\bf 0.54} & 21.37& 21.63 & 65.22 & \underline{0.69} & 55.92 & 6.69 \\
Blended(Random)      & 75.98 & 99.91  & 59.78 & \underline{0.34} & 63.34 & 12.39 &23.77 &67.56 & 57.28 & {\bf 0.08} & 57.26 & 22.88 \\
      
      WaNet & 74.11 & 96.03 & 60.98 & 57.25 & 63.12 & \underline{39.18} &20.60 &43.10 & 63.16 & 80.59 & 53.46 & {\bf 5.94}  \\
    \midrule
    Average &75.89  & 97.85 & 60.29 & 43.90 & 61.80 & \underline{20.80} &23.24 &48.00 & 64.32 & 45.09 & 56.18 & {\bf 5.46} \\
    \bottomrule
  \end{tabular}
  }
  \vskip -0.05in
  \label{tab:main_cifar100}
\end{table*}




 \begin{figure*}[!htbp]
 \centering
 \subfigure[BadNets($3\times 3$)]{
 \label{fig:abla_badnets}
 \includegraphics[scale=0.18]{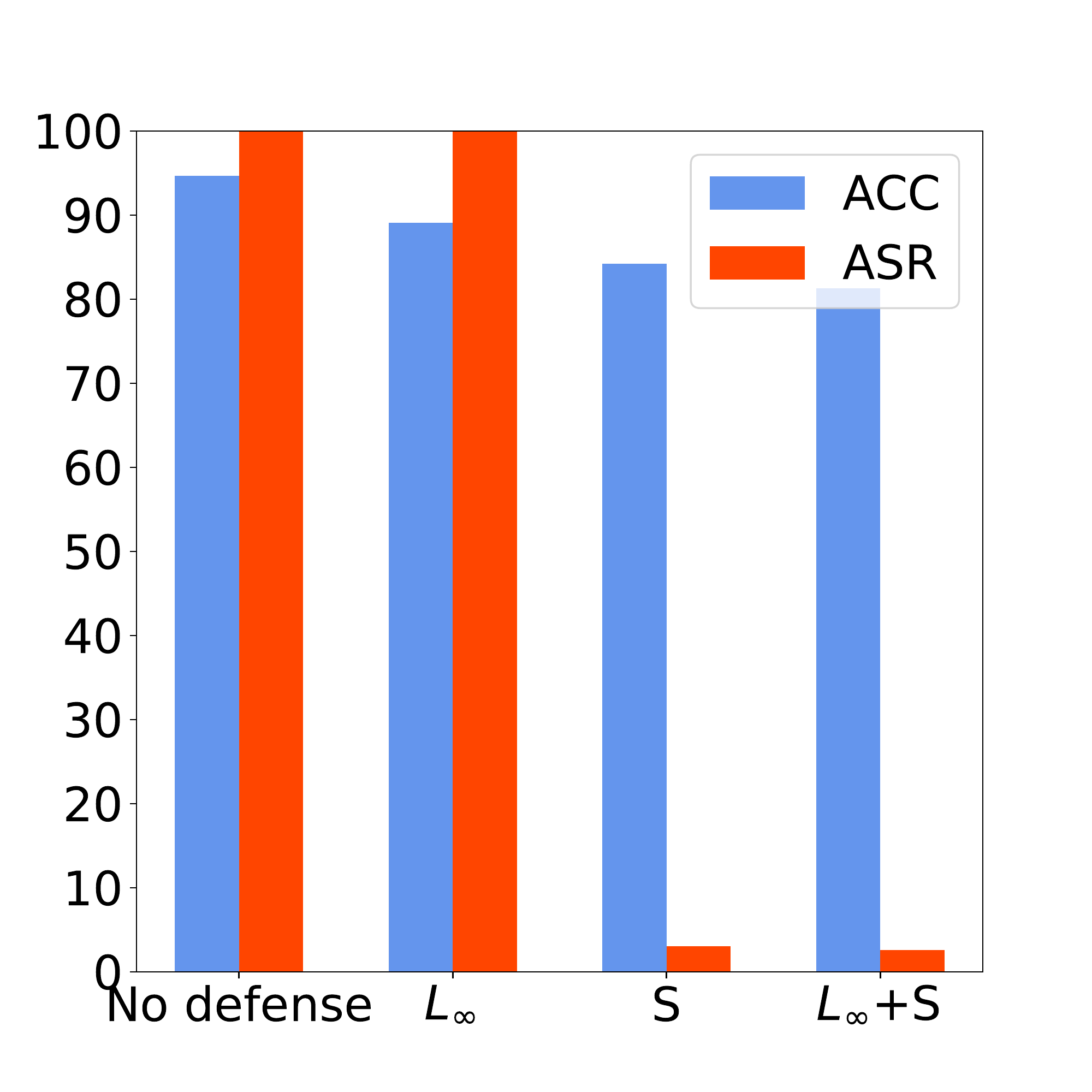}}
 \subfigure[LC]{
 \label{fig:abla_lc}
 \includegraphics[scale=0.18]{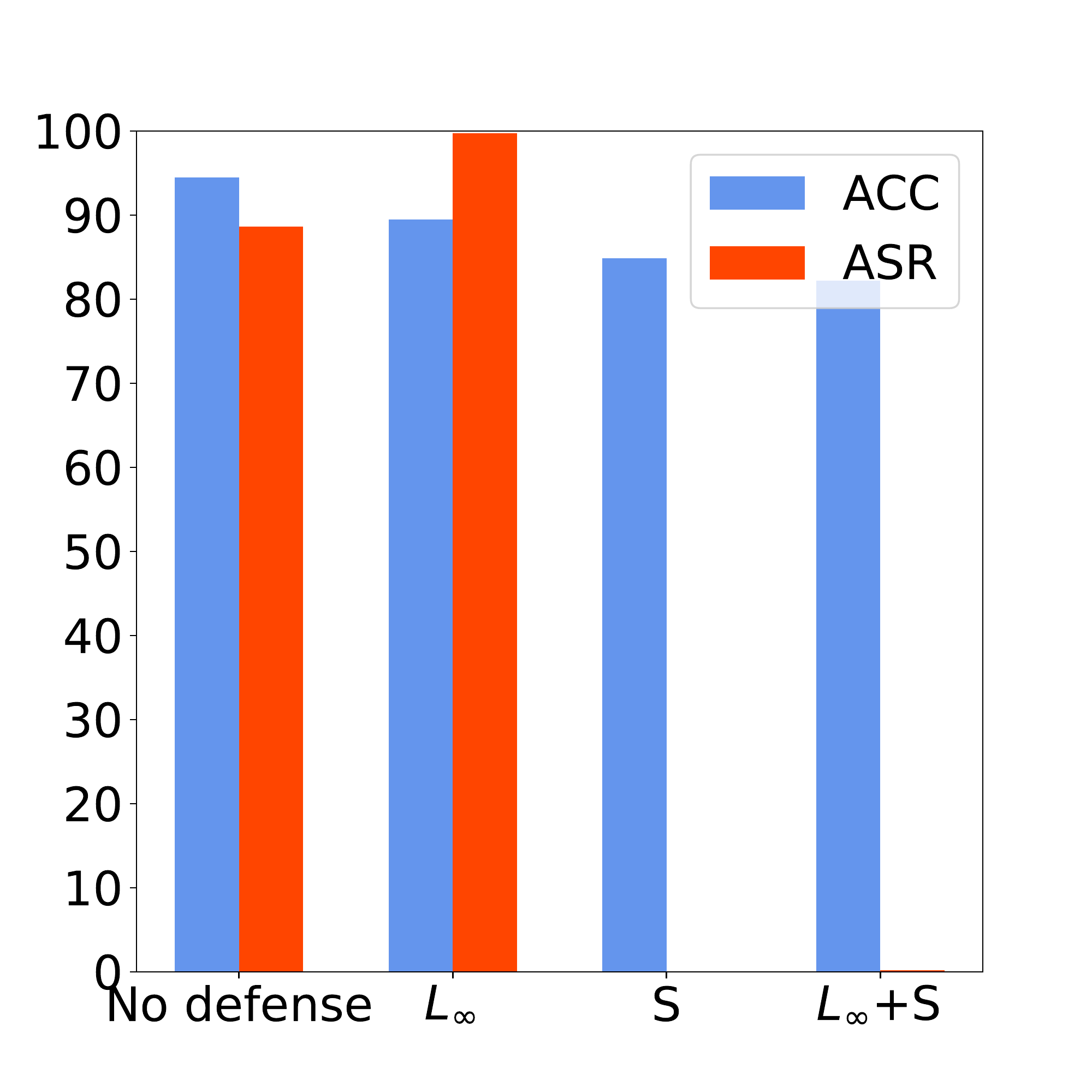}}
 \subfigure[Blended(Hello-Kitty)]{
 \label{fig:abla_blended}
 \includegraphics[scale=0.18]{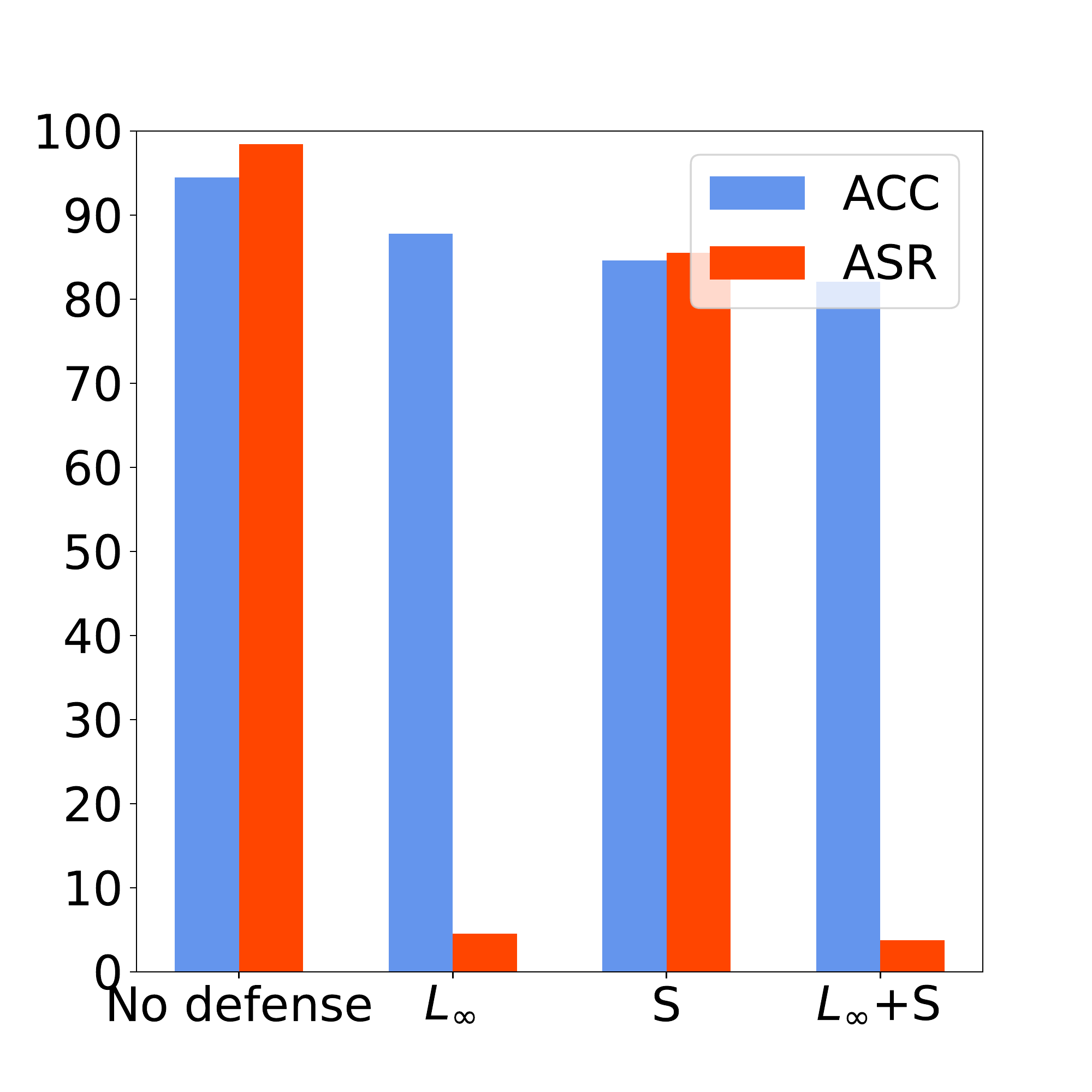}}
 \subfigure[WaNet]{
 \label{fig:abla_wanet}
 \includegraphics[scale=0.18]{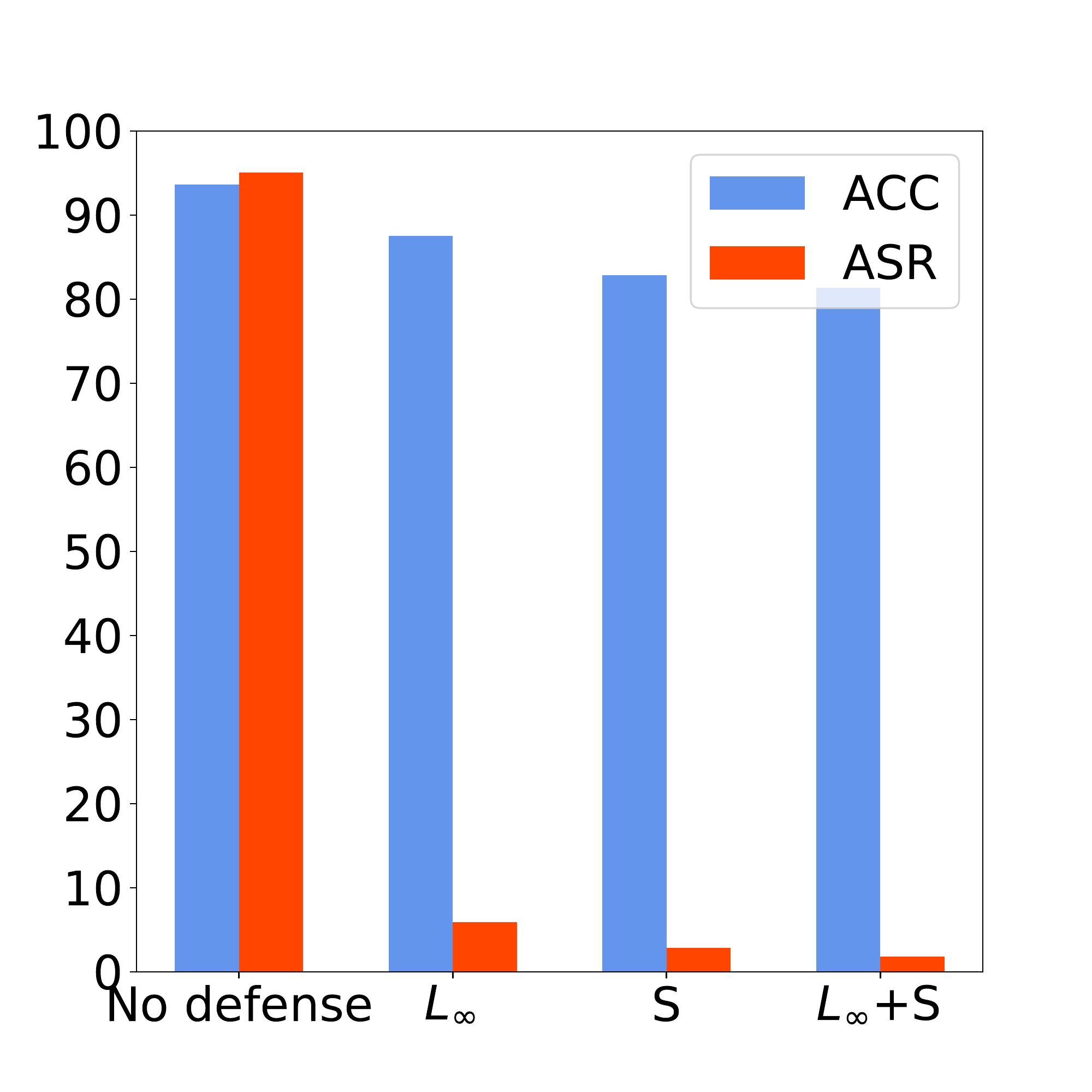}}
 \caption{Comparison with a single adversarial perturbation. `S' is the abbreviation of `Spatial'.  }
 \label{fig:ablation}
 \end{figure*}

\begin{figure*}[!htbp]
\centering
\subfigure[BadNets($3\times 3$)]{
\label{fig:other_badnets}
\includegraphics[scale=0.18]{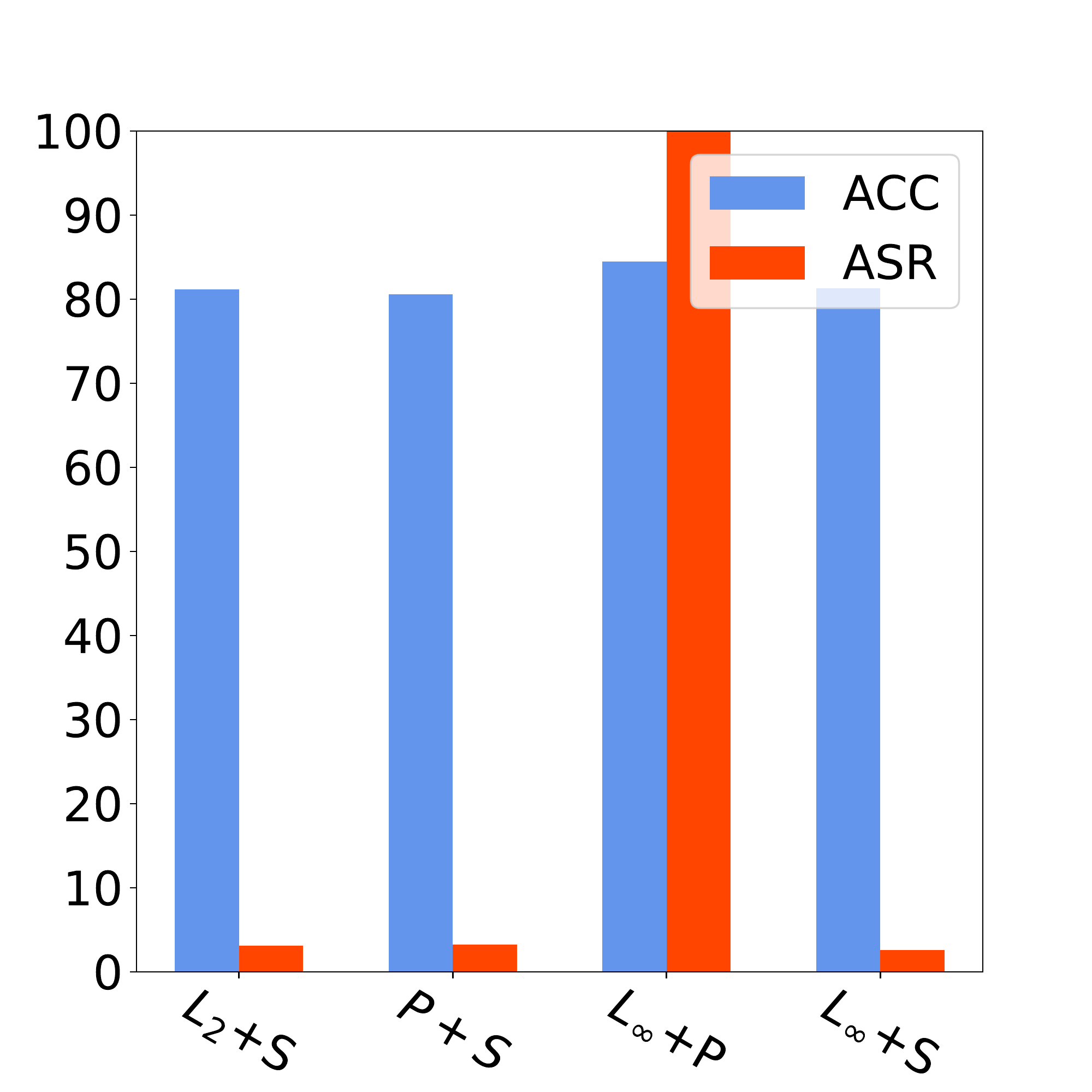}}
\subfigure[LC]{
\label{fig:other_lc}
\includegraphics[scale=0.18]{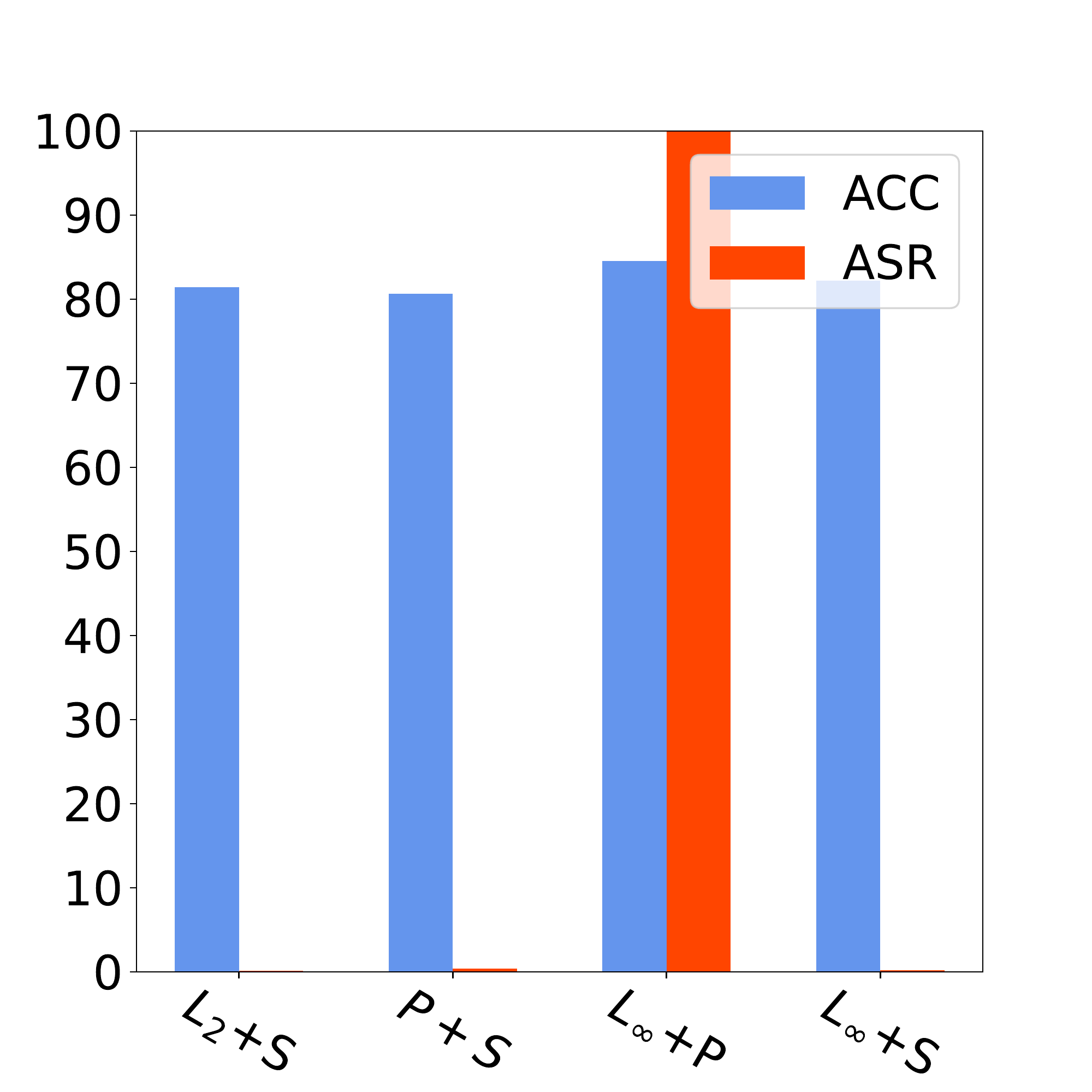}}
\subfigure[Blended(Hello-Kitty)]{
\label{fig:other_blended}
\includegraphics[scale=0.18]{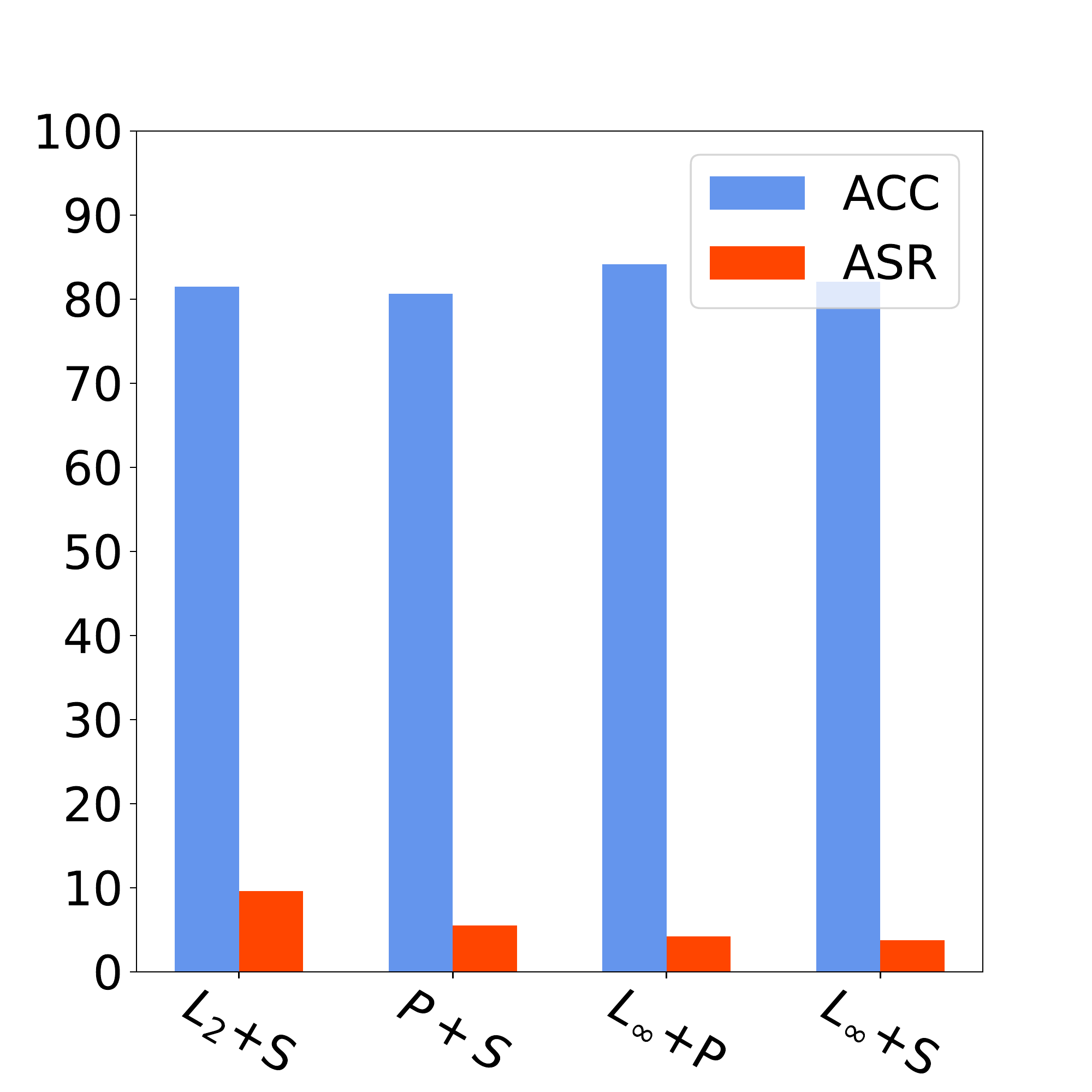}}
\subfigure[WaNet]{
\label{fig:other_wanet}
\includegraphics[scale=0.18]{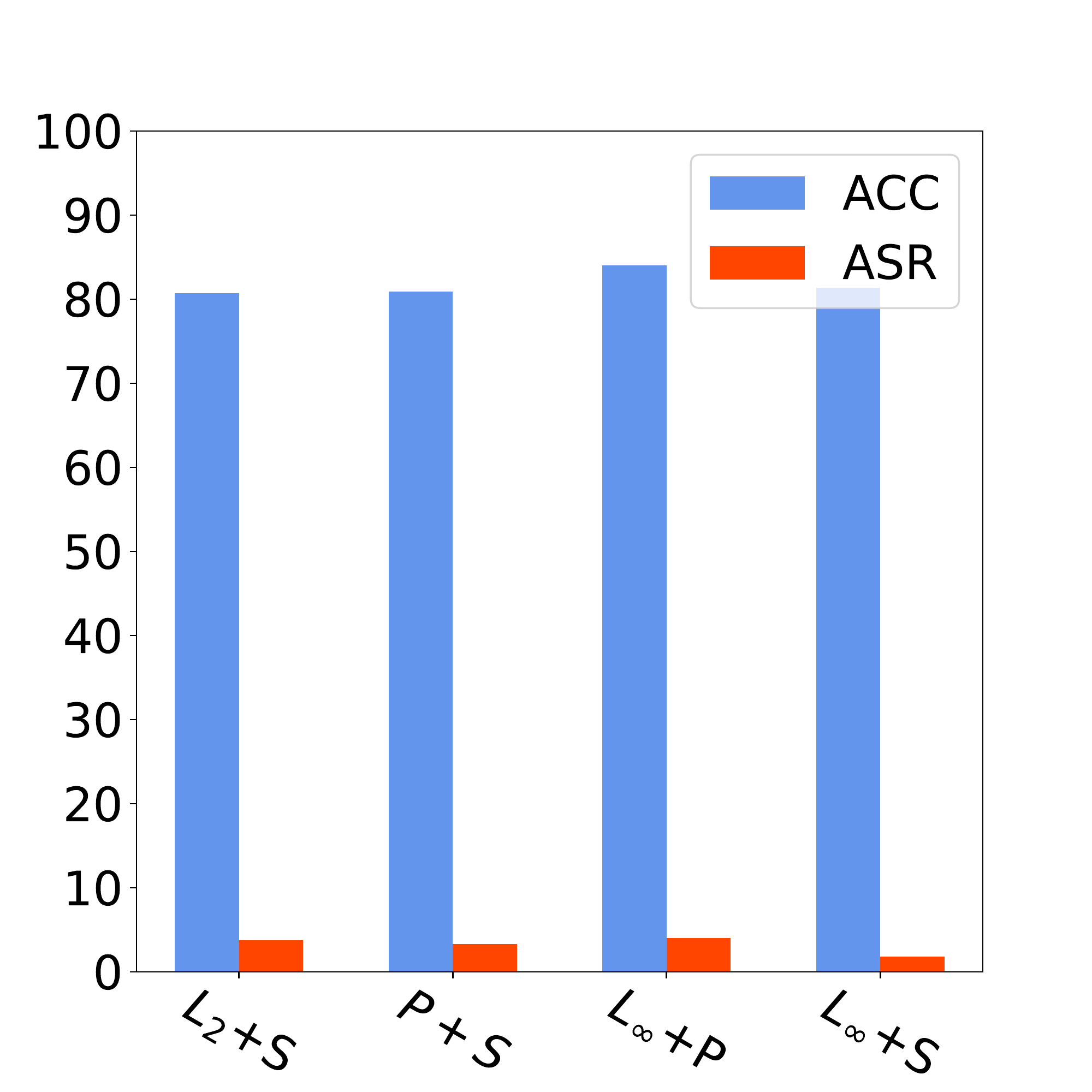}}
\caption{Comparison with other combinations of AT. `S' is the abbreviation of `Spatial'. `P' is the abbreviation of `Perceptual'.  }
\label{fig:other_cat}
\end{figure*}


{\bf Main Results.} As shown in Table \ref{tab:main_cifar10}, we find that CAT outperforms the baseline methods in most cases, which demonstrates the effectiveness of the composite strategy. Compared with FP and NAD, a major advantage is that, CAT doed not need extra clean data, which leads to wider applications as clean data may be hard to collect in some area.  Compared with DPSGD and ABL, CAT achieves more stable and better results in the terms of backdoor robustness. For CIFAR-10, the lowest ASR of ABL is $0.00\%$ (LC) while the highest ASR is $98.83\%$ (BadNets ($2\times 2$)). The lowest ASR of DPSGD is $4.13\%$ (BadNets (Random)) while the highest ASR is $99.92\%$
(Blended (Random)).
However, CAT decreases all attacks' ASRs below $6\%$. CAT also achieves more stable results for CIFAR-100. We attribute the results to the difference of the technical strategy. Although the three methods (DPSGD, ABL and CAT) aim to train clean models with poisoned data from scratch,  ABL identifies the candidates of poisoned data in the early training stage and forgets them later. However, the inaccuracy of detecting the poisoned data and the gradient ascent used in ABL tend to cause the training instability, which will not happen in CAT as we only perturb the training data with  imperceptible noise.  DPSGD perturbs gradients with noise to minimize the difference between clean gradients and poisoned ones. The perturbation leads to a significant drop in the clean accuracy, and yet does not provide meaningful guarantees. One limitation of CAT is that the adversarial perturbations lead to the clean accuracy drop, which is universal in AT methods, and we leave the improvements for our future work.

\section{Discussion}

In this section, we mainly discuss the ablation experiments, other combinations of AT and the potential adaptive attack.

{\bf The Necessity of Combination in CAT.}  We conducted an extra experiment with only one single adversarial perturbation. In Figure \ref{fig:ablation}, we find that AT with a single perturbation is not able to cope with all types of backdoor attacks, while CAT does. For example, although $L_\infty$ AT successfully mitigates Blended Attacks, it is ineffective against BadNets. While Spatial AT easily defends against BadNets, it can only decreases the ASR of Blended Attacks to $\thicksim 80\%$. 

{\bf Other Combinations of AT.} We also considered other combinations in CAT: `$L_2$
+ Spatial', `Perceptual + Spatial', `$L_\infty$ + Perceptual'. The results are summarized in Figure \ref{fig:other_cat}, from which we find that  `$L_2$
+ Spatial', `Perceptual + Spatial' and `$L_\infty$+Spatial' make similar effects on the above four types of backdoor attacks. Although `$L_\infty$ + Perceptual' successfully mitigates Blended Attacks and WaNet, it does not work for BadNets and LC since $L_\infty$ and perceptual adversarial perturbations can not effectively destroy the patch form.


{\bf Adaptive Attack.} We consider the scenario that the adversary is aware of our AT method and propose an adaptive attack. We formulate the adaptive attack as a bi-level optimization problem. The core is that we aim to learn specific perturbations that can induce the adversarially trained model to misclassify triggered samples. Gradient matching \citep{geiping2020witches, zhao2020dataset} and model retraining techniques \citep{souri2021sleeper} were utilized to solve adaptive attack better. However, in our experiments, we did not succeed to break AT. We conjecture that this is due to the high non-convexity of the bi-level optimization.  We leave the exact formulations and experimental results in Appendix \ref{app:adapt}.  



\section{Conclusion and Future Work}

In this work, we conducted thorough experiments to investigate the effects of AT on backdoor attacks. Our results suggested that prior findings may ignore the influences of the perturbation budget, the threat model used in AT and the trigger forms in backdoor attacks. AT actually showed the effectiveness of mitigating backdoor attacks in many cases. We further proposed composite AT to address unknown backdoor attacks. Through extensive experiments, we demonstrated that CAT outperforms other baseline methods. We believe that our work sheds lights on understanding the interactions between AT and backdoor attacks and encourages the researchers to evaluate the effectiveness of backdoor attacks in the AT framework.  

For future work, we will continue to improve AT from several aspects. One direction is to explore more suitable adversarial perturbations against backdoor attacks than the four types used in this paper. We have found that spatial transformation perturbation efficiently mitigates patch-based attacks and $L_p$
adversarial perturbation efficiently addresses whole-image attacks. However, whether there exist more suitable adversarial perturbation remains unknown. Besides, a major limitation is the clean accuracy drop using AT and another interesting direction is to improve the natural generalization of AT and maintain backdoor robustness in the meantime, which may be achieved by borrowing the techniques in adversarial learning community. 

\bibliography{templateArxiv.bib}

\bibliographystyle{unsrt}  


\clearpage

\appendix
\section{Experimental Settings for CIFAR-100 }
\label{app:cifar100}



\textbf{Experimental Settings in Section \ref{section-revisit}.} For CIFAR-100, we evaluated four backdoor attacks: BadNets with a $3\times 3$ checkerboard trigger, LC, Blended with a Hello-Kitty trigger, and WaNet. We adopted the clean label setting for BadNets while three other attacks were implemented based on the original papers. The poison rate  was $0.3\%$ for BadNets and LC, $3\%$ for Blended and $5\%$ for WaNet. For all attacks, class 2 was assigned as the target class. 

\textbf{Experimental Settings in Section \ref{section-cat}.} For CIFAR-100, the poison rate  was $0.5\%$ for BadNets and LC, $3\%$ for Blended and $5\%$ WaNet. For all attacks, class 2 was assigned as the target class. 

\section{Adaptive Attack}
\label{app:adapt}
\textbf{Bi-level Optimization.} We formulate the adaptive attack as a bi-level optimization problem:
\begin{equation}
\begin{aligned}
    \min\limits_{\delta} &\sum_{i=1}^n \ell(f_{\theta}(x_i+t), y_t) \\
    \text{s.t.} \ \ \theta = \arg\min_{\tau} &\sum_{i=1}^n \max_{z_i\in \mathcal{B}(x_i+\delta_i)} \ell(f_\tau(z_i), y_i),\\
    \Vert \delta \Vert \le \delta_{\max},
\label{equ:adapt}
\end{aligned}
\end{equation}
where $t$
denotes the trigger (\textit{e.g.} a patch),
$y_t$
denotes the target class.
The difference between the above objective and Souri \textit{et al.} \citep{souri2021sleeper} is that we use the adversarial data rather than original data in the lower-level problem since our goal is to design the adaptive attack against AT. We could interpret the objective function from two perspectives: \textit{the upper-level optimization} and \textit{the lower-level optimization}. The lower-level optimization obtains the optimal model parameters with AT. The upper-level optimization aims to minimize the loss on backdoor images. The final goal is to acquire the optimal perturbations $\delta$ which can mislead the adversarially trained model to classify the backdoor images as target class. We solve the bi-level optimization with a surrogate objective \citep{souri2021sleeper}:
\[ \mathcal{A}=1- \frac{\nabla_\theta \ell(\theta)\cdot \nabla_\tau \ell(\tau) }{\Vert\ell(\theta)\cdot \ell(\tau)\Vert}, \]
which is named as gradient matching \citep{geiping2020witches}. We also use the model retraining and poison selection techniques suggested in \citep{souri2021sleeper}. The whole procedure is similar with \citep{souri2021sleeper} except that we use adversarial retraining rather than standard retraining (see Algorithm \ref{alg1}).

\textbf{Experimental Settings.} In our experiments, we set the optimization steps $R=250$ and retraining factor $T=4$.
The poison rate was $1\%$. We used the same $\epsilon$ values when optimizing the perturbation $\delta$ in Equation (\ref{equ:adapt}) and training the victim model. We conducted adaptive attack with various $\delta_{\max}$ values and trained the poisoned data with $\epsilon=8/255 \  L_\infty$ AT and standard training ($\epsilon=0$). The results are summarized in Table \ref{tab:adapt_attack}, from which we find that although the adaptive attack successfully breaks standard training, it does not work for $L_\infty$ AT whenever $\delta_{\max}$ is larger than $\epsilon$ or not. We conjecture that such results are due to the high convexity of the bi-level optimization.

\begin{algorithm}
 \caption{Adaptive Attack}
 \begin{algorithmic}[1]
 \renewcommand{\algorithmicrequire}{\textbf{Input:}}
 \renewcommand{\algorithmicensure}{\textbf{Begin:}}
 \REQUIRE Surrogate network $f_\tau$
 , training data $\mathcal{D} = \{(x_i, y_i)\}_{i=1}^n$
 , trigger $t$
 ,target label $y_t$, poison budget
 $m \leq n$
 , adversarial budget $\epsilon$,
 optimization steps $R$
 , retraining factor $T$
    \STATE Select $m$ samples with label  $y_t$ from $\mathcal{D}$
    with the highest gradient norms. 
    \STATE Randomly initialize perturbations $\{\delta_i\}_{i=1}^m$
    \FOR {$r$ = 1,\ 2,\ ...\ ,\ $R$ optimizations steps}
        \STATE Compute $ \mathcal{A}$ and update $\{\delta_i\}_{i=1}^m $ with a step of signed Adam.
        \IF {$r \mod  \lfloor{R/(T+1)}\rfloor  = 0 $ and $r \ne R$}
            \STATE Retrain $f$ on poisoned training data with AT\\ $\{(x_i + \delta_i, y_i)\}_{i=1}^m \cup \{(x_i, y_i)\}_{i=m+1}^n $ and \\ update $\tau$
    \ENDIF
    \ENDFOR
 \RETURN  $\{\delta_i\}_{i=1}^m$ 
 \end{algorithmic} 
 \label{alg1}
 \end{algorithm}

\begin{table}[!htbp]
\caption{Results of adaptive attack. }
\begin{center}
\scalebox{0.9}{
\begin{tabular}{c|cc}
\toprule
 & ACC & ASR  \\ 
\midrule
Adaptive attack ($\delta_{\max}=4/255, \epsilon=0$) &  94.43 &1.66 \\
Adaptive attack ($\delta_{\max}=8/255, \epsilon=0$) &  93.92 &25.10 \\
Adaptive attack ($\delta_{\max}=16/255, \epsilon=0$) &  94.50 &48.68 \\
\midrule
Adaptive attack ($\delta_{\max}=4/255, \epsilon=8/255$) &  84.51 & 2.79\\
Adaptive attack ($\delta_{\max}=8/255, \epsilon=8/255$) &  84.70 & 2.67\\
Adaptive attack ($\delta_{\max}=16/255, \epsilon=8/255$) & 84.76  & 2.49\\
\bottomrule
\end{tabular}
}
\end{center}
\label{tab:adapt_attack}
\end{table}


\end{document}